\pdfoutput=1

\documentclass[a4paper,fleqn]{cas-sc}
\usepackage{multirow}
\usepackage{algorithm}
\usepackage[noend]{algpseudocode}
\usepackage[utf8]{inputenc}
\usepackage[T1]{fontenc}
\usepackage{textcomp}
\usepackage{subfig}
\usepackage[title]{appendix}
\usepackage[authoryear]{natbib}
\usepackage{lineno}
\usepackage[doublespacing]{setspace}
\usepackage{url}
\usepackage{breakurl}

\let\oldequation\equation
\let\oldendequation\endequation

\renewenvironment{equation}
  {\linenomathNonumbers\oldequation}
  {\oldendequation\endlinenomath}

\begin{document}
\let\WriteBookmarks\relax
\def\floatpagepagefraction{1}
\def\textpagefraction{.001}
\shorttitle{}
\shortauthors{Mouret et~al.}

\title [mode = title]{Reconstruction of Sentinel-2 Derived Time Series Using Robust Gaussian Mixture Models \textemdash Application to the Detection of Anomalous Crop Development}                      
\tnotemark[1]

\tnotetext[1]{This document is the results of the research project funded by TerraNIS SAS. and ANRT (convention CIFRE no. 2018/1349). It was partly supported Minciencias through the 910-2019-Program ECOS Nord entitled "Fusion of synthetic aperture radar and optical multitemporal imaging and its application to anomaly detection in citrus" under Grant 8598.}
\tnotetext[2]{This is a preprint version of the research paper published in Elsevier Computers and Electronics in Agriculture: \url{https://doi.org/10.1016/j.compag.2022.106983}.}

\author[1,2]{Florian Mouret}[type=editor,
                        orcid=0000-0003-3719-6767 ,bioid=1,
                        ]
\cormark[1]
\ead{florian.mouret@terranis.fr / florian.mouret@irit.fr}

\author[1]{Mohanad Albughdadi}
\author[1]{Sylvie Duthoit}
\author[3]{Denis Kouamé}
\author[1]{Guillaume Rieu}
\author[2]{Jean-Yves Tourneret}

\address[1]{TerraNIS, 12 Avenue de l'Europe, 31520 Ramonville-Saint-Agne, France}
\address[2]{University of Toulouse / IRIT-INP-ENSEEIHT / TéSA, 2 Rue Charles Camichel, 31000 Toulouse, France}
\address[3]{University of Toulouse / IRIT-UPS, 118 Route de Narbonne, 31062 Toulouse Cedex 9, France}

\cortext[cor1]{Corresponding author}
\begin{abstract}
Missing data is a recurrent problem in remote sensing, mainly due to cloud coverage for multispectral images and acquisition problems. This can be a critical issue for crop monitoring, especially for applications relying on machine learning techniques, which generally assume that the feature matrix does not have missing values. This paper proposes a Gaussian Mixture Model (GMM) for the reconstruction of parcel-level features extracted from multispectral images. A robust version of the GMM is also investigated, since datasets can be contaminated by inaccurate samples or features (e.g., wrong crop type reported, inaccurate boundaries, undetected clouds, etc). Additional features extracted from Synthetic Aperture Radar (SAR) images using Sentinel-1 data are also used to provide complementary information and improve the imputations. The robust GMM investigated in this work assigns reduced weights to the outliers during the estimation of the GMM parameters, which improves the final reconstruction. These weights are computed at each step of an Expectation-Maximization (EM) algorithm by using outlier scores provided by the isolation forest (IF) algorithm. Experimental validation is conducted on rapeseed and wheat parcels located in the Beauce region (France). Overall, we show that the GMM imputation method outperforms other reconstruction strategies. A mean absolute error (MAE) of $0.013$ (resp. $0.019$) is obtained for the imputation of the median Normalized Difference Index (NDVI) of the rapeseed (resp. wheat) parcels. Other indicators (e.g., Normalized Difference Water Index) and statistics (for instance the interquartile range, which captures heterogeneity among the parcel indicator) are reconstructed at the same time with good accuracy. In a dataset contaminated by irrelevant samples, using the robust GMM is recommended since the standard GMM imputation can lead to inaccurate imputed values. An application to the monitoring of anomalous crop development in the presence of missing data is finally considered. In this application, using the proposed method leads to the best detection results, especially when SAR data are used jointly with multispectral images. Exploiting the information contained in cloudy multispectral images instead of removing these images is beneficial for this application. \end{abstract}




\maketitle
\section{Introduction}\label{sec:intro}

Remote sensing images have become an essential tool for many agricultural applications, including precision farming \cite{WEISS2020111402}, primarily because they can be used to provide valuable information about vegetation without a need for on-site visits \citep{SCHULZ2021106173}. In recent years, the amount of freely accessible remote sensed images has drastically increased, especially thanks to the Copernicus mission operated by the European Space Agency (ESA). Its first multispectral high resolution satellite (Sentinel-2A) was launched in 2015, followed by a second satellite in 2017 (Sentinel-2B) \citep{DRUSCH201225}. Two synthetic aperture radar (SAR) satellites, Sentinel-1A and Sentinel-1B, are also part of the Copernicus mission and were launched in 2014 and 2016 \citep{TORRES20129}. Sentinel-1 (S1) and Sentinel-2 (S2) images are available with high temporal and spatial resolutions, which is well suited for precision agriculture. S2 images have been widely used for crop mapping and monitoring, \textit{e.g.}, for the detection of land use anomalies \citep{Kanjir_2018}, the estimation of biophysical parameters such as the leaf area index (LAI) \citep{VERRELST2015260, ALBUGHDADI2021105899} or more generally to provide information on the vegetation status \citep{DEFOURNY2019551}. Since S1 images can provide information regarding water content and the structure of the vegetation \citep{Khabbazan2019}, they have been also largely investigated for crop monitoring and mapping \citep{Vreugdenhil2018, MANSARAY2020105674, NASIRZADEHDIZAJI2021106118}. Finally, the joint use of these sensors has been motivated by their complementary \citep{Inglada_2016, Navarro_2016, VELOSO2017415, Denize2018, Mouret_2021}.

\begin{figure}[htp!]
    \centering
    \includegraphics[width=0.5\textwidth]{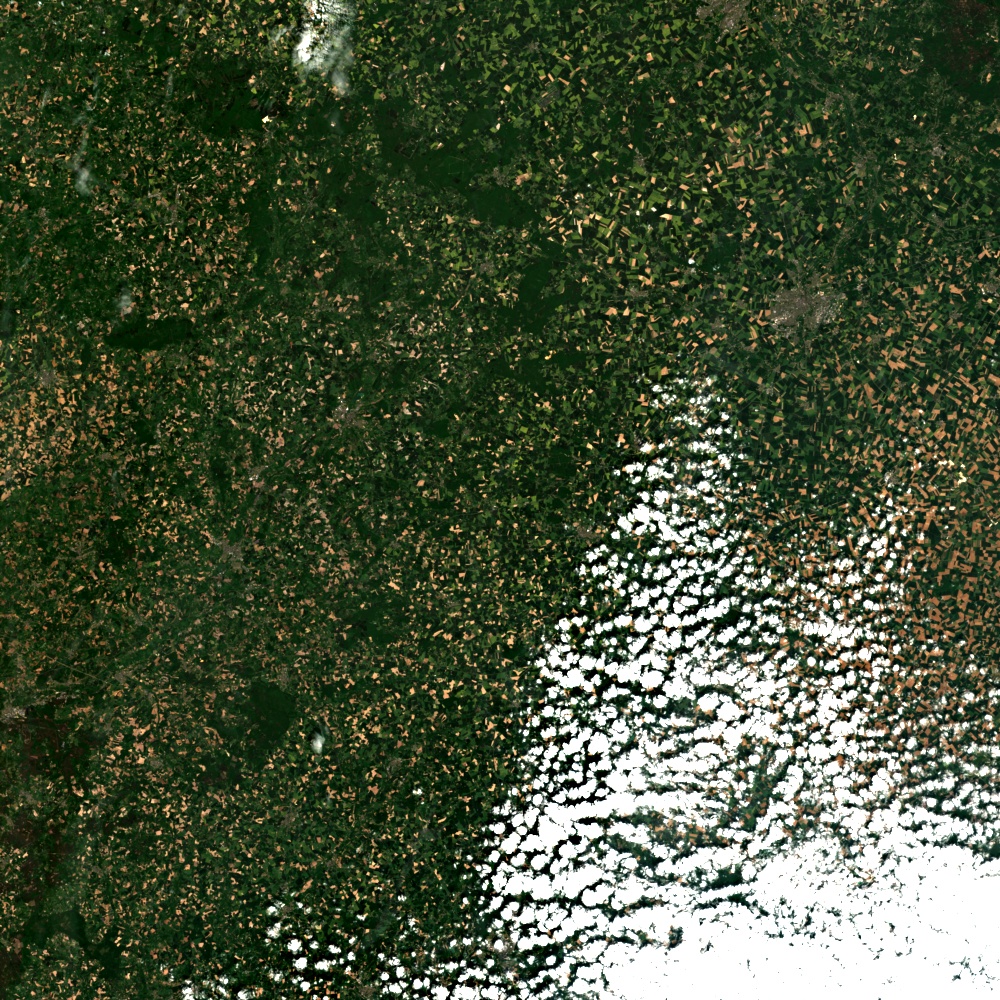}
    \caption{S2 image (natural color) acquired in May 09 2018 (tile T31UCP analyzed in this paper). A part of the image is covered by clouds (bottom right)}
    \label{fig:ex_cloud}
\end{figure}

A main challenge that can affect all the aforementioned applications is the presence of missing data, which is an inherent problem in remote sensing. Multispectral images are particularly sensitive to this issue since they are affected by clouds (to a lesser extent, acquisition problems can also affect SAR images). An illustrative example is provided in \autoref{fig:ex_cloud}, where part of a S2 image is covered by clouds. The problem of missing data is of crucial importance when using machine learning techniques, which generally assume a complete feature matrix. The lack of timely information on crops has been identified for decades as a main limitation for precision agriculture based on remote sensing \citep{MORAN1997319}. This paper focuses on the reconstruction of multiple parcel-level features extracted from S1 and S2 data, when part of S2 images are missing due to the presence of clouds. The features used in this study were previously investigated for the detection of abnormal crop development at the parcel level \citep{Mouret_2021}, after discarding the parcels with missing values, which is obviously not acceptable for operational applications. To that extent, we also propose to evaluate the interest of considering partially cloudy images (with missing data) for this application.

Various methods have been proposed in the remote sensing literature to deal with missing data. A general review \citep{Shen2015} has grouped the different methods into four categories: 1) spatial-based methods, 2) spectral-based methods, 3) temporal-based methods and 4) hybrid methods (combining the spatial, spectral and temporal strategies). For crop monitoring, temporal-based and hybrid methods are generally used, since the temporal information is an essential indicator when analyzing the vegetation status. Temporal-based methods are also known as ``gap filling'' and traditionally rely on linear or spline interpolations. They are well suited to dense noisy time series and have provided interesting results, \textit{e.g.}, for the classification of crop types or the prediction of plant diversity \citep{Inglada_2015, VUOLO2017202, FAUVEL2020111536}. However, they can lack precision when there is a need to monitor abrupt changes or when data from a large period of time is missing. Hybrid methods have been used intensively in remote sensing, mostly because they are able to impute missing data in multimodal signals and images, such as multispectral and SAR images. Recent techniques based on deep learning have also been investigated for SAR-Optical image matching \citep{Mazza, Hughes}. Image matching can be interesting to reconstruct large parts of S2 image. However it generally uses a single SAR image acquired at a date close to the multispectral image to be reconstructed. Consequently, this method does not fully exploit all the available data acquired throughout the growing season. To address this issue, some methods have been designed to reconstruct S2 images based on multi-temporal SAR/Optical data \citep{Bermudez_2019, Ebel_2022}. However, these methods are not suited to our use case for various reasons: 1) they require a supervised learning step whereas the proposed strategy is fully unsupervised, avoiding any costly labelling, 2) they focus on the reconstruction of full S2 images while in this study we focus on parcel-level statistics of vegetation indices and 3) they necessitate a huge amount of training data (for example, \cite{Ebel_2022} uses a training set composed of 40 ROI of size $4000\times4000$ $px^2$). Deep learning methods have also been used to regress NDVI time series based on SAR times series and various other external indicators (e.g., weather, terrain) \citep{Garioud_2020, Garioud_2021}. However, while being relevant to impute dense time series for large scale applications, these methods need a huge amount of training data (more than $23 850$ parcels are analyzed in \cite{Garioud_2020} and even more in \cite{Garioud_2021}), which is not always accessible in practice. For instance, the French Land Parcel Identification System (LPIS) used in these studies is generally available with a delay of one or two years. Moreover, the method proposed by \citet{Garioud_2021} has been designed to express NDVI as a function of SAR time series and does not exploit the available S2 information for the imputation task. Similarly, \citet{PIPIA2019111452} proposed to estimate the leaf area index (LAI) at the pixel level using Gaussian processes. Nevertheless, this method has been designed to reconstruct a single optical time series using a single SAR time series, which is too restrictive for the problem addressed here. Finally, interesting results have been obtained with methods ignoring missing data induced by clouds for specific applications, such as land cover classification \citep{russwurm_2017, Castro_2020}.  However, this paper focuses on a more generic use case, since having access to reconstructed time series can be interesting for a wider range of applications.

The method investigated in this paper can impute missing features derived from S2 data in a robust fashion by using a Gaussian Mixture Model (GMM) \citep{DEMP1977} learned using the Expectation-Maximization (EM) algorithm. The main originality of the proposed approach is to use outlier scores resulting from an outlier detection algorithm within the EM algorithm to 1) detect abnormal agricultural parcels and 2) have a robust parameter estimation of the GMM parameters. GMMs have been used successfully in remote sensing, \textit{e.g.}, for clustering \citep{Lopes2017} and supervised classification \citep{Tadjudin_2000, Lagrange2017}. However, despite their natural ability to reconstruct missing data \citep{Ghahramani_1994, EIROLA201432}, they have not been investigated for crop monitoring (to the best of our knowledge). The main motivation for using GMMs is their faculty to learn complex behaviors in a fully unsupervised way. Even if these models also suffer from the curse of dimensionality, they can be used with a limited amount of data, which is important here since the number of analyzed parcels is relatively small (the database used in the experiments contains around $2000$ parcels).

The rest of this paper is organized as follows.  Section~\ref{sec:method} presents the study area with the available data and the proposed method. Experimental results are presented in Section~\ref{sec:results}, with an application to the detection of anomalies in the development of rapeseed crops (additional experiments were conducted on wheat crops and are available in the supplementary materials). In Section~\ref{sec:discussion}, a discussion on the results and the different imputation methods is proposed. Finally, some conclusions are drawn in Section~\ref{sec:conclusion}.

\section{Materials and Methods}\label{sec:method}

\subsection{Data}\label{sec:studyarea}

\subsubsection{Study Area and Parcels}
\begin{figure}[htp!]
    \centering
    \includegraphics[width=1\textwidth]{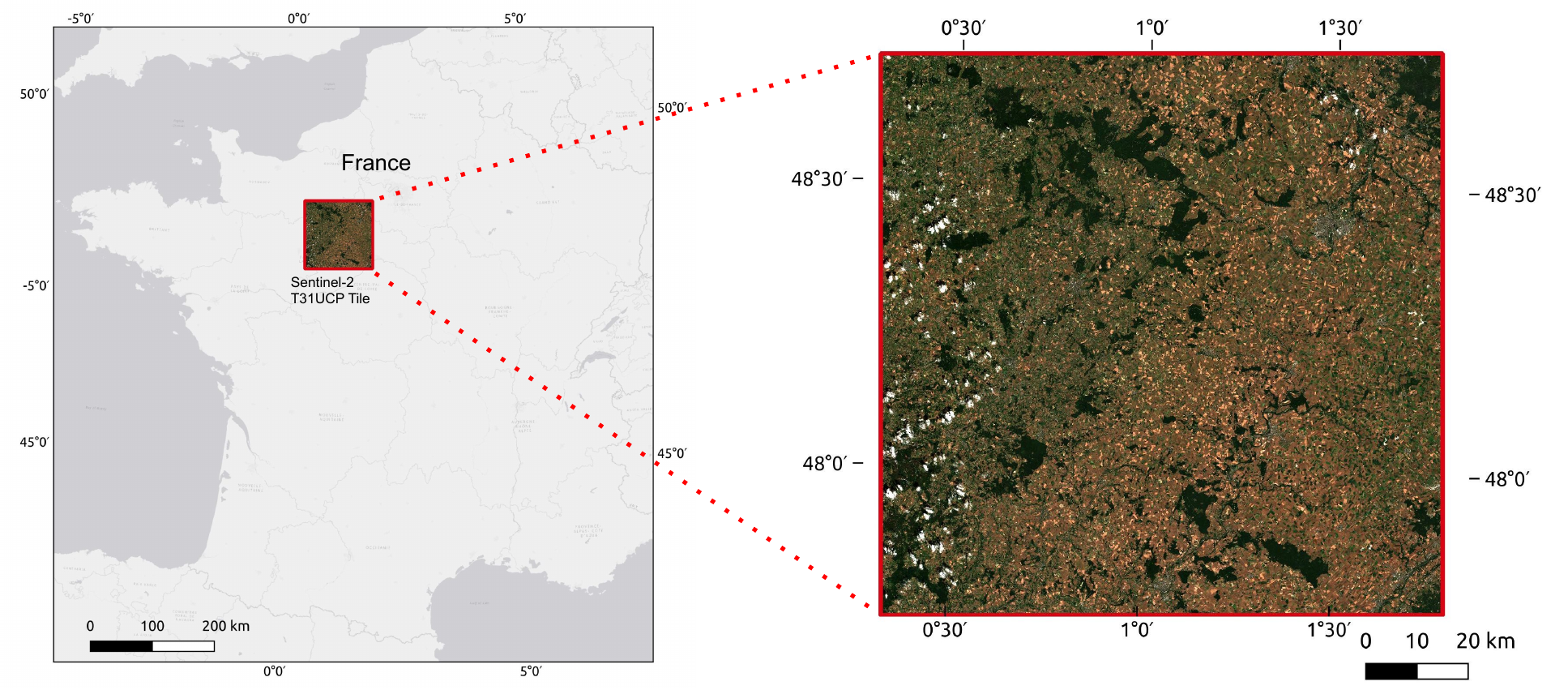}
    \caption{The Sentinel-2 tile considered in this work is delimited by the red box. The S2 image processed in level 2A acquired in June 28 2018 is displayed in natural colors in the right part of the Figure.}
    \label{fig:zone_etude}
\end{figure}

The area of interest is located in France (Beauce region) and corresponds to the S2 tile T31UCP (whose center is located approximately at 48°24’N latitude and 1°00’E longitude) as depicted in \autoref{fig:zone_etude}. In this study area, $2218$ rapeseed parcels are monitored for the 2017/2018 growing seasons. All the parcels affected by clouds were discarded in a first analysis in order to have a reliable ground-truth to validate the proposed imputation method. Note that experiments were also conducted on $3361$ wheat parcels (for the growing season 2016/2017). Since the results obtained confirmed those obtained on rapeseed parcels, they are reported in the supplementary material attached to this document for the sake of conciseness.

\subsubsection{Satellite Data}
S1-A and S1-B satellites are SAR C-band imaging satellites whose center frequency is $5.405$ GHz. The data collected for this study has been acquired in both ascending and descending orbit passes and in dual polarization (VV and VH). Ground Range Detected (GRD) products are used in the Interferometric Wide (IW) swath mode (phase information is lost but the volume of data is drastically reduced) \citep{TORRES20129}. The following preprocessing steps have been applied to each S1 image: thermal noise removal, calibration, terrain flattening and range Doppler terrain correction. Multi-temporal speckle filtering was tested, but without bringing any improvement for the considered application. Overall, $40$ S1 images were selected for the growing season analyzed. Note that a drop in the number of available S1 images was observed after April 2018 (this can be observed in all web pages related to sentinel data, confirming an acquisition problem).

\begin{figure}[ht]
\centering
\includegraphics[width=0.5\textwidth]{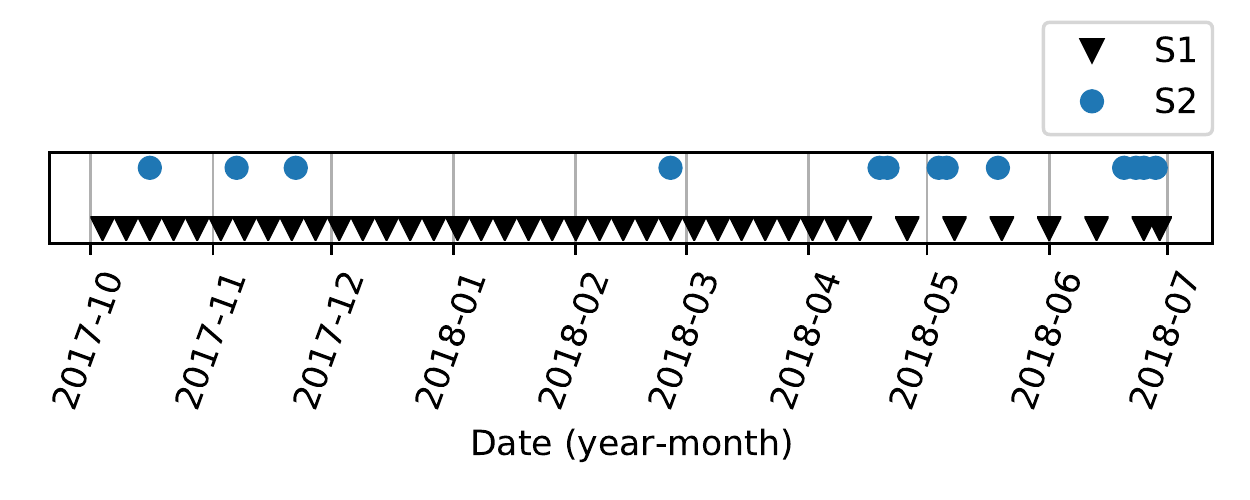}
\caption{Each marker corresponds to the acquisition date of a used image for the growing season 2017/2018.}
\label{fig:calendar}
\end{figure}

S2-A and S2-B satellites are multispectral imaging satellites with $13$ spectral bands covering the visible, the near infra-red (NIR) and the shortwave-infrared (SWIR) spectral region \citep{DRUSCH201225}. Level-2A ortho-rectified products expressed in surface reflectance are obtained using the MAJA processing chain \citep{Hagolle2015}, which also provides cloud and shadow masks. Overall, $13$ S2 images were selected for the considered growing season. This low number of images is due to the cloud coverage, which often leads to images fully covered by clouds, especially in winter. For instance, no S2 image is exploitable between December 2017 and the end of February 2018, as shown in \autoref{fig:calendar}.


\subsubsection{Pixel-level features}

The features extracted at the pixel-level from S1 and S2 images are detailed in \autoref{table:table_optical_indicators}. Each of these features showed interest for crop monitoring and the detection of anomalous development, as detailed in \citep{Mouret_2021}. S2 features consist of 5 benchmark agronomic vegetation indices (VI), namely the Normalized Difference Vegetation Index (NDVI) \citep{rouse1974monitoring}, the Green-Red Vegetation Index (GRVI) \citep{Motohka2010}, two variants of the Normalized Difference Water Index (NDWI) \citep{Gao1996, McFEETERS1996} and a variant of the Modified Chlorophyll Absorption Ratio Index (MCARI/OSAVI) \citep{DAUGHTRY2000, Wu2008}. These 5 VI use different spectral bands combinations, each one focusing on a particular spectral response of the vegetation. S1 features are directly the VV and VH backscattering coefficients. Combinations of these coefficients were tested (e.g., their ratio) but without improving the quality of the detection results \citep{Mouret_2021}. Note that other features could be added or removed (depending on the application) without changes in the proposed imputation approach.

\begin{table}[htp!]
\caption{Pixel-level features extracted from S2 and S1 images used for the crop analysis. For S2, the near infrared (band~8), red edge (band~5), short wave infrared (band~11), green (band~3) and red (band~4) channels are denoted as NIR, RE, SWIR, GREEN and RED, respectively \citep{Mouret_2021}. S2 indicators are in the range [-1,1], except for MCARI/OSAVI varying in the range $[0, +\infty]$.}
\centering
\begin{tabular}{lll}
\\\hline Sensor type              & Indicator & Expression  \\ \hline & \\
\multirow{10}{*}{Multispectral} & \textnormal{NDVI}      & $(\textnormal{NIR} - \textnormal{RED}) / (\textnormal{NIR} + \textnormal{RED})$ \\
                         & $\textnormal{NDWI}_{\textnormal{SWIR}}$     & $(\textnormal{NIR} - \textnormal{SWIR}) / (\textnormal{NIR} + \textnormal{SWIR})$     \\
                         & $\textnormal{NDWI}_{\textnormal{GREEN}}$      & $(\textnormal{GREEN} - \textnormal{NIR})/(\textnormal{GREEN} + \textnormal{NIR})$     \\
                         & $\textnormal{MCARI}/\textnormal{OSAVI}$      & $\left[(\textnormal{RE} - \textnormal{IR}) - 0.2(\textnormal{RE} - \textnormal{RED})\right] /        \left[(1+0.16)\frac{\textnormal{NIR} - \textnormal{RED}}{\textnormal{NIR} + \textnormal{RED} + 0.16}\right]$         \\
                         & \textnormal{GRVI}      & $(\textnormal{GREEN} - \textnormal{RED})/(\textnormal{GREEN} + \textnormal{RED})$      \\ & \\ \hline & \\ \multirow{4}{*}{SAR}     & \makecell[l]{Cross-polarized backscattering \\ coefficient VH}        & $\gamma^0_{\textnormal{VH}}$       \\ & \\
                         & \makecell[l]{Co-polarized backscattering \\ coefficient VV}        & $\gamma^0_{\textnormal{VV}}$\\
                        \\\hline
\end{tabular}
\label{table:table_optical_indicators}
\end{table}

\subsubsection{Parcel-level features and creation of the feature matrix}\label{sec:features}

\begin{figure}[htp!]
    \centering
    \includegraphics[width=1\textwidth]{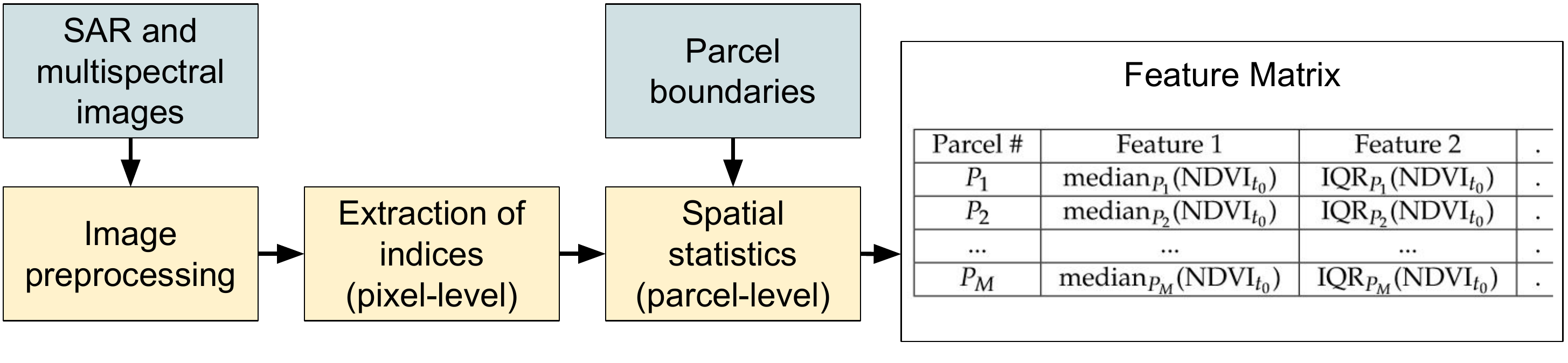}
    \caption{The different steps leading to the creation of the feature matrix used for the detection of parcels with anomalous behavior. NDVI$_{t_n}$ denotes the NDVI computed at time instant $t_n$ while median$_{P_M}$ and $\text{IQR}_{P_M}$ denote the spatial median and IQR for the indicator computed inside the parcel $\# M$.}
    \label{fig:workflow}
\end{figure}

Parcel-level features are computed from pixel-level features by using two spatial statistics: the median and the interquartile range (IQR). The median is a robust mean value of an indicator, and can capture the mean behavior of a given parcel. The IQR captures the dispersion of an indicator, providing relevant information at the parcel-level regarding potential heterogeneous development of the vegetation. Note that IQR is not computed for the S1 features, since it is directly proportional to the median. Each parcel is characterized by the different parcel-level statistics computed at each acquisition. The number of columns of the feature matrix is $N_{c}=  N_{1,i} \times N_{1,f} \times N_{1,s} + N_{2,i} \times N_{2,f} \times N_{2,s} $, where $N_{1,i}$ is the number of S1 images, $N_{1,f}$ is the number of pixel-level features extracted for each S1 image, $N_{1,s}$ is the number of statistics computed for each S1 feature. Similar definitions apply to $N_{2,i}$, $N_{2,f}$ and $N_{2,s}$ for S2 images. The processing chain that leads to the creation of the feature matrix used for the detection of outlier parcels is summarized in \autoref{fig:workflow}. Each line of the feature matrix has $210$ features characterizing each rapeseed parcel (when considering all the acquisition dates and all the spatial statistics, i.e., $130$ S2 features and $80$ S1 features).

Note that the study area and the data were also used in \citep{Mouret_2021}. The interested reader is invited to consult this reference for complementary information and additional examples, especially regarding the description of the parcels with anomalous development.

\subsection{Imputation of Missing Values with Mixture of Gaussians using the EM Algorithm}

The proposed approach uses a multivariate GMM to impute the potential missing values of the feature matrix. The GMM parameters can be estimated in a fully unsupervised manner, using the Expectation-Maximization (EM) algorithm. Within the EM algorithm, maximum likelihood (ML) estimation of the model parameters is conducted. This estimation can be naturally conducted in the presence of missing data \citep{DEMP1977, Ghahramani_1994}. In that case, the ML estimation is evaluated on the observed values and missing values are estimated using the current model parameters. The resulting probabilistic model is very flexible and can be applied to data of relative small size. After presenting the general principle of the EM algorithm for GMM estimation, we introduce a robust modification of this method taking into account the presence of outliers in the dataset and improving the estimation of the model parameters. More details about GMMs can be found in the classic book from \citet{Bishop2006}, while an interesting review dealing with regularization techniques for GMM in high dimension was proposed in \citet{BOUVEYRON_review}. Regarding the different application of GMMs for agricultural machine vision systems, it is worth mentioning the review proposed by \citet{REHMAN2019585}. An interesting application of the EM algorithm for the detection of cucumber disease was considered in \citet{ZHANG2017338}. In what follows, we only provide key results for the estimation of GMM with the EM algorithm in the presence of missing data. More precisely, Section\ref{sec:general_EM} recalls the principles of the EM algorithm for imputing missing data using GMM, which has been studied for instance in \cite{DEMP1977, Ghahramani_1994}. Section\ref{sec:robust_gmm} explains how to robustify this algorithm to mitigate the presence of outliers, which is the main contribution of this work. Detailed derivations and theoretical justifications are provided in Appendix~\ref{appendix:standard_EM}. The algorithm used for the proposed robust EM for GMM is also provided in Appendix~\ref{appendix:algo}.

\subsubsection{Imputation of missing data using the EM algorithm for GMM}\label{sec:general_EM}

In what follows, we suppose that each of the $N$ samples of the feature matrix $X$ is distributed according to a mixture of $K$ Gaussian distributions. In the presence of missing values, each sample can be decomposed into $\boldsymbol{x}_n = (\boldsymbol{x}_n^{o_n}, \boldsymbol{x}_n^{m_n})$, where $\boldsymbol{x}_n^{o_n}$ and $\boldsymbol{x}_n^{m_n}$ are the vectors of observed and missing variables respectively. More generally, the superscripts $o_n$ and $m_n$ denote the observed and missing components of the sample $n$. These subscripts can be used for matrices too, \textit{e.g.}, $\boldsymbol{\Sigma}_k^{o_n m_n}$ refers to the elements of the matrix $\boldsymbol{\Sigma}_k$ in the rows and columns specified by $o_n$ and $m_n$ (and so on). For brevity, we will denote $o_n = o$ and $m_n = m$. The EM algorithm aims at maximizing the observed (or complete) log-likelihood $\log \mathcal{L}_c(\boldsymbol{\theta};\boldsymbol{X}^o, \boldsymbol{X}^m, \boldsymbol{z})$ with respect to the model parameters:\begin{equation}
    \log \mathcal{L}_c(\boldsymbol{\theta};\boldsymbol{X}^o, \boldsymbol{X}^m, \boldsymbol{z}) = \sum_{n=1}^{N} \sum_{k=1}^{K} z_{nk} \log \left[ \pi_k \mathcal{N}(\boldsymbol{x}_n^o | \boldsymbol{\mu}_k, \boldsymbol{\Sigma}_k) \right],
\end{equation}
with $\boldsymbol{X}^o$ the set of all observed variables, $\boldsymbol{X}^m$ the set of all missing variables, $\mathcal{N}(\boldsymbol{x}_n^o | \boldsymbol{\mu}_k, \boldsymbol{\Sigma}_k)$ the marginal multivariate normal probability density of the observed sample $\boldsymbol{x}_n^o$ and $\boldsymbol{\theta} =$ $ \{\pi_1, ..., \pi_k, $ $\boldsymbol{\mu}_1, ..., \boldsymbol{\mu}_k, \boldsymbol{\Sigma}_1, ..., \boldsymbol{\Sigma}_K \}$ the set of parameters to be estimated. These parameters are the mean vectors $\boldsymbol{\mu}_k$, the covariance matrices $\boldsymbol{\Sigma}_k$ and the mixing coefficients $\pi_k$ ($0 < \pi_k < 1, \sum_{k=1}^{K}\pi_k = 1$) of the GMM. 

The E-step evaluates $E[\log \mathcal{L}_c(\boldsymbol{\theta};\boldsymbol{X}^o, \boldsymbol{X}^m, \boldsymbol{z})| \boldsymbol{\theta}, \boldsymbol{x}^o]$. In practice, it reduces to compute the following sufficient statistics:
\begin{linenomath}
\begin{align}
    & \gamma_{nk} = \frac{\pi_k \mathcal{N}(\boldsymbol{x}_n^o, \boldsymbol{\mu}_k^o, \boldsymbol{\Sigma}_k^{oo})}{\sum_{j=1}^{K} \pi_j \mathcal{N}(\boldsymbol{x}_n^o, \boldsymbol{\mu}_j^o, \boldsymbol{\Sigma}_j^{oo})},\\
    & \hat{\boldsymbol{\mu}}_{nk}^m = \boldsymbol{\mu}_k^m + \boldsymbol{\Sigma}_k^{mo}(\boldsymbol{\Sigma}_k^{oo})^{-1}(\boldsymbol{x}_n^o - \boldsymbol{\mu}_k^o),
    \label{eq:conditional_mean}\\
    & \hat{\boldsymbol{x}}_{nk}^m = (\boldsymbol{x}_n^o, \boldsymbol{\mu}_{nk}^m),
    \label{eq:x_hat}\\
    & \hat{\boldsymbol{\Sigma}}_{nk}^{mm} = \boldsymbol{\Sigma}_k^{mm} - \boldsymbol{\Sigma}_k^{mo}(\boldsymbol{\Sigma}_k^{oo})^{-1}\boldsymbol{\Sigma}_k^{mo},
    \label{eq:conditional_cov}\\
    & \hat{\boldsymbol{\Sigma}}_{nk} = \begin{pmatrix} 
    \boldsymbol{0}^{oo} & \boldsymbol{0}^{om} \\
    \boldsymbol{0}^{mo} & \hat{\boldsymbol{\Sigma}}_{nk}^{mm} \\
    \end{pmatrix},
    \label{eq:sigma_hat}
\end{align}
\end{linenomath}
where $\boldsymbol{0}^{oo}, \boldsymbol{0}^{om}$ and $\boldsymbol{0}^{mo}$ are matrices of zeros of appropriate dimensions.

The responsibility $\gamma_{nk}$ is the probability that samples $n$ has been generated by the $k$-th component. More precisely, $\gamma_{nk} = E[z_{ik} | \boldsymbol{\theta}^{(t)}, \boldsymbol{x}^o]$ (this is similar to the complete-data case except that $\gamma_{nk}$ is evaluated on the observed part of the data). The other terms are specific to the missing data case. They results from the evaluation of $E\left[(\boldsymbol{x}_n - \boldsymbol{\mu}_k)\boldsymbol{\Sigma}_k^{-1}(\boldsymbol{x}_n - \boldsymbol{\mu}_k) | \boldsymbol{\theta}, \boldsymbol{x}_n^o \right]$. Note that~\eqref{eq:conditional_mean} and \eqref{eq:conditional_cov} are the conditional expectation and the conditional covariance matrix of the missing variables of a sample given that $\boldsymbol{x}_n$ has been generated by Gaussian $\# k$, \textit{i.e.}, $\hat{\boldsymbol{\mu}}_{nk}^m = E[\boldsymbol{x}_n^m | \boldsymbol{x}_n^o]$ and $\hat{\boldsymbol{\Sigma}}_{nk}^{mm} = \textrm{Var}[\boldsymbol{x}_n^m | \boldsymbol{x}_n^o]$. Note also that the missing values of $x_{nk}$ have been replaced by their expectations $\hat{\boldsymbol{\mu}}_{nk}^m$ in~\eqref{eq:x_hat}. Similarly, the matrix $\hat{\boldsymbol{\Sigma}}_{nk}$ of~\eqref{eq:sigma_hat} has been filled with zeros except for the missing components, which leads to $\hat{\boldsymbol{\Sigma}}_{nk}^{mm}$.

In the M-step, maximizing the current expectation lead to the following new set of parameters:
\begin{equation}
    \boldsymbol{\mu}_k^{(t+1)} = \frac{1}{N_k}\sum_{n=1}^{N} \gamma_{nk} \hat{\boldsymbol{x}}_n, \; \boldsymbol{\Sigma}_k^{(t+1)} = \frac{1}{N_k}\sum_{n=1}^{N}\gamma_{nk}\left[(\hat{\boldsymbol{x}}_n - \hat{\boldsymbol{\mu}}_k)(\hat{\boldsymbol{x}}_n - \hat{\boldsymbol{\mu}}_k)^T + \hat{\boldsymbol{\Sigma}}_{nk} \right], \; \pi_k^{(t+1)} = \frac{N_k}{N},
\end{equation}
where with $N_k = \sum_{n=1}^{N} \gamma_{nk}$. These expressions are similar to the complete-data case, except that missing values have been imputed using the expression obtained in the E-step and covariance matrices are corrected to take into account the presence of missing values. More details about the EM algorithm for the GMM with missing data can be found for instance in \citet{Ghahramani_1994} and it is worth mentioning the interesting work conducted in \citet{EIROLA201432} devoted to the estimation of distances with missing values and applied to various tasks including classification and regression.

Finally, note that when estimating GMMs, a regularization of the covariance matrices is generally needed to avoid instabilities and numerical issues. The strategy adopted for our implementation is based on the method proposed by \cite{BOUVEYRON_article}. It consists in setting the smallest eigenvalues of the covariance matrices to a constant. More details on that point are available in Appendix~\ref{sec:regularization}.

\subsubsection{Robust GMM}\label{sec:robust_gmm}

\begin{figure}[htp]
    \centering
    \includegraphics[width=0.5\textwidth]{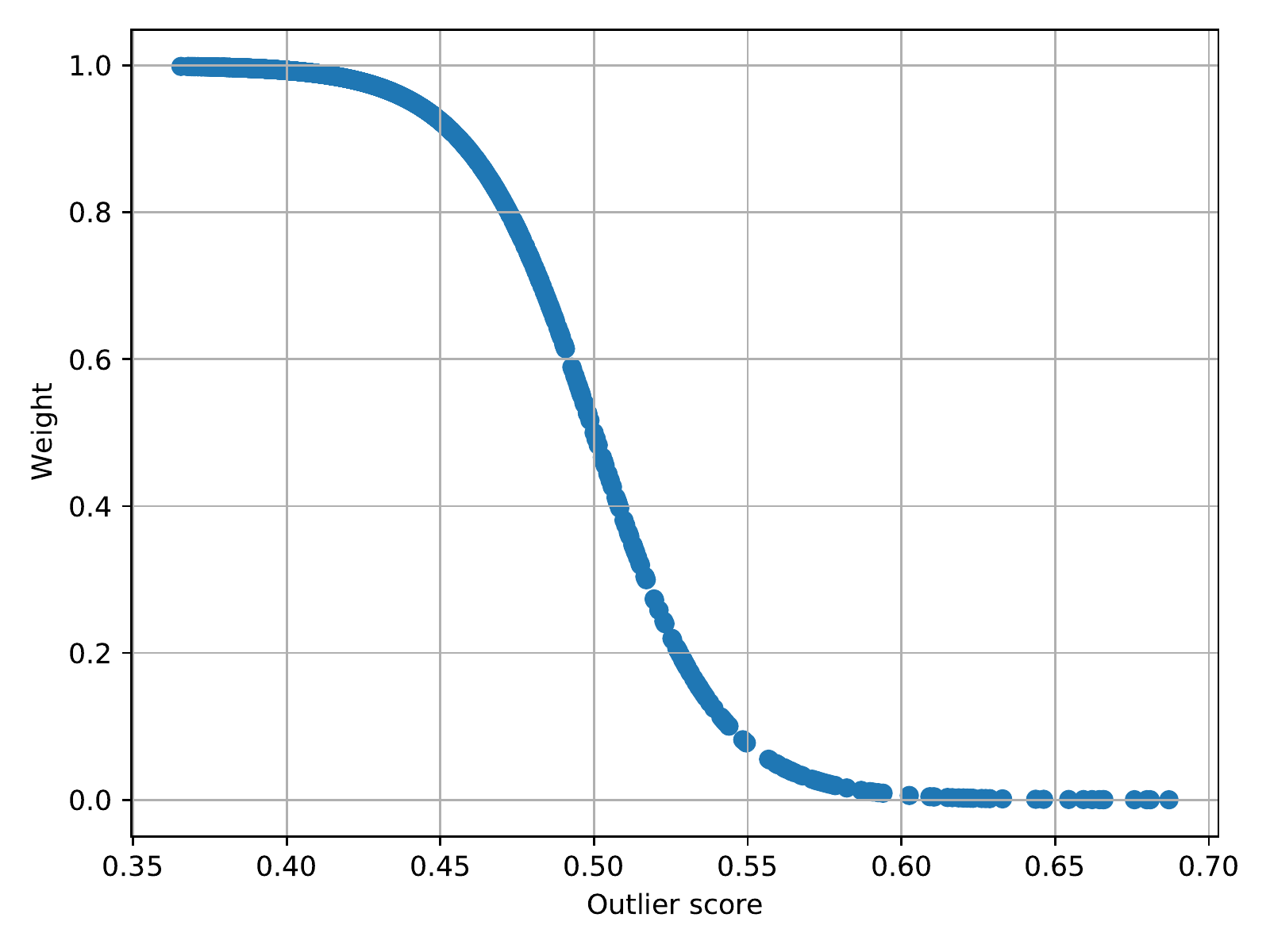}
    \caption{Variation of the weight $w_n$ versus the outlier score attributed by the IF algorithm, with $\alpha=50$ and $\text{th}=0.5$.}
    \label{fig:repartition}
\end{figure}

The estimation of the means and covariances of a GMM using the EM algorithm is sensitive to the presence of outliers, especially in the M-step \citep{Campbell_1984, Tadjudin_2000}. To address this issue in the context of semi supervised classification with remote sensing images, \citet{Tadjudin_2000} introduced a robust parameter estimation method which associates weights to the observed samples. The idea is that samples with a reduced weight (corresponding ideally to outliers) will have a small influence on the estimation of the model parameters. However, the method proposed in \citet{Tadjudin_2000} suffers from two main limitations: 1) It does not detect the outliers in an unsupervised way and 2) It does not take into account the presence of missing data. To overcome these issues, we propose to modify this method by using the output of the Isolation Forest (IF) algorithm, which is a reference method for the detection of the outliers \citep{Liu2012}. This algorithm was found to be efficient to detect relevant  abnormal parcels \citep{Mouret_2021} and has the advantage of providing an outlier score in the range [0,1]. In order to build a robust GMM, we propose to weight the importance of each sample in the M-step by using the anomaly score provided by the IF algorithm. The resulting robust EM algorithm updates the unknown GMM parameters in the M-step as in \cite{Tadjudin_2000}
\begin{equation}\label{eq:robust_mstep}
    \boldsymbol{\mu}_{k}^{(t+1)} = \frac{\sum_{n=1}^{N} w_n \gamma_{nk} \hat{\boldsymbol{x}}_{nk}}{\sum_{n=1}^{N} w_n \gamma_{nk}} , \boldsymbol{\Sigma}_k^{(t+1)} = \frac{\sum_{i=1}^{N} w_n^2 \gamma_{nk}\left[(\hat{\boldsymbol{x}}_n - \hat{\boldsymbol{\mu}}_k)(\hat{\boldsymbol{x}}_n - \hat{\boldsymbol{\mu}}_k)^T + \hat{\boldsymbol{\Sigma}}_{nk} \right]}{\sum_{n=1}^{N} w_n^2 \gamma_{nk}}, \pi_k^{(t+1)} = \frac{N_k}{N}.
\end{equation}
However, contrary to \cite{Tadjudin_2000}, the weights $w_n$ are computed using the outlier score of the IF algorithm (denoted as $\textrm{score}_{\textrm{IF}}(\boldsymbol{\hat{\boldsymbol{x}}}_n)$ for the imputed sample $\boldsymbol{\hat{\boldsymbol{x}}}_n$) as follows:
\begin{equation}\label{eq:weight_if}
    w_{n} = \frac{1}{1 + \exp{[\alpha(\textrm{score}_{\textrm{IF}}(\boldsymbol{\hat{\boldsymbol{x}}}_n) - \text{th})}]},
\end{equation}
where $\alpha$ and $\text{th}$ are two constants to be fixed by the user. Motivations for using the sigmoid \eqref{eq:weight_if} include the fact that it is a smooth monotonically function of the weights taking its values in the range [0,1], with a unique inflection point equal to $\text{th}$. Note that the parameter $\alpha$ controls the speed of the inflection: for high values of $\text{th}$, the sigmoid \eqref{eq:weight_if} reduces to a hard thresholding operation around $\text{th}$, whereas it decreases more slowly from $1$ to $0$ for lower values of $\text{th}$. A score of $0.5$ is a natural threshold in the IF algorithm, as explained in \citet{Liu2012}. An example of the evolution of the weights with respect to their anomaly score is depicted in \autoref{fig:repartition} for $\alpha=50$ and $\text{th}=0.5$. 

\subsection{Evaluation method and simulation scenarios}

This section provides details on the simulation scenarios, which were designed to evaluate the presence of missing values in the observed data. We also introduce metrics to evaluate the performance of the different algorithms considered to reconstruct the missing data. An application to the detection of anomalies in agricultural parcels is finally described with appropriate performance measures.

\subsubsection{Simulation scenarios}
In order to evaluate the performance of missing data reconstruction with a controlled ground truth, we removed some existing features in the dataset introduced in Section \ref{sec:features}. Two parameters control the number of missing data: the percentage of S2 images having missing values (\textit{e.g.}, due to the presence of clouds), and for each of these S2 images, the percentage of parcels affected by missing values (the parcels affected by clouds are not necessarily the same for each S2 image). For a given S2 image with missing data, we have removed all the features associated with this image for all the affected parcels. In practice, for cloudy days, missing values are likely to affect a significant amount of the parcels. In the presented experiments, half of the total number of parcels (chosen equally likely in the database) was supposed to be cloudy with all S2 features removed  (other tests were made with different percentages of cloudy images leading to similar conclusions). The different scenarios considered in this paper are summarized in \autoref{table:experiments}.

\begin{table}[pos=htp]
\caption{Summary of the experiments conducted in this paper with the percentage of cloudy S2 images and the percentage of cloudy parcels within a given cloudy S2 image. The column ``Cloudy S2 images'' indicates the percentage of S2 images with missing values whereas the column ``Affected parcels'' provides the percentage of parcels with missing values within a cloudy S2 image.}
\footnotesize
\begin{tabular}{llll}
Section & Evaluated factor                                                 & Cloudy S2 images & Affected parcels  \\\hline
\ref{appendix:convergence}     & Convergence of the EM algorithm                                  & 8\% (1 S2 image)                          & 50\%                                                                   \\
\ref{sec:results_rapeseed}   & Effect of the percentage of S2 images affected by missing values & Varies in {[}0, 70{]}\%          & 50\%                                                                   \\
\ref{sec:contamination_experiment}   & Effect of adding irrelevant samples                              & 23\% (3 S2 images)                        & 50\%                                                                   \\
\ref{sec:result_application}    & Application to the detection of abnormal crop development        & Varies in {[}0, 70{]}\%          & 50\%    
\end{tabular}
\label{table:experiments}
\end{table}

\subsubsection{Performance measures}\label{sec:metrics}

The mean absolute reconstruction error (MAE) is used to evaluate the quality of the reconstruction of the different missing features, with the advantage of being unambiguous and naturally understandable compared to the root mean squared error (RMSE) \citep{Willmott}. The MAE is defined as follows:
\begin{equation}
    \textrm{MAE} = \frac{\sum_{i=1}^{N_m} | f_{i} - \hat{f}_{i}| }{N_{m}},
\end{equation}
where $N_{m}$ is the number of missing features, $f_i$ is the original value of the $i$th feature and $\hat{f}_{i}$ denotes its estimation (also referred to as imputation or reconstruction).

The second set of experiments shows the usefulness of the proposed method for the detection of abnormal crop development in the presence of missing data. Using an outlier detection algorithm (here, the IF algorithm) to detect potential anomalies, the precision of the results (defined by the number of true positives divided by the total number of parcels detected as outliers) is computed for various outlier ratios. This outlier ratio is fixed by the user and corresponds to the percentage of abnormal parcels to be detected. The area under the precision vs. outlier ratio curve (AUC) is a good metric to measure the ability of an outlier detection algorithm to detect relevant anomalies. It is similar to the precision vs. recall curves or receiver operational characteristics with the advantage of being adapted to imbalanced datasets containing outliers \citep{Saito_2015, Mouret_2021}.

\section{Results}\label{sec:results}

Both robust and non-robust GMM imputation methods are compared to the imputation obtained using the k-nearest neighbors (KNN)\citep{Troyanskaya2001}. Note that the non-robust version of the GMM is regularized using the technique mentioned in Section~\ref{sec:regularization}. Various other imputation methods (gap filling, autoencoders, multiple imputations, soft imputation) were tested and are discussed in Section~\ref{sec:discussion}. They are not presented here for conciseness since they did not improve our results. The results presented in this paper focus on the imputation of multispectral S2 time series, but the same method could be used to reconstruct S1 features as well. Finally, note that before the GMM and KNN imputations, each feature was scaled in the range $[0,1]$ using the available values, \textit{i.e.}, without the missing data (features in natural scale can then be obtained using the inverse transformation). The hyperparameters used for the different reconstruction algorithms are reported in \autoref{table:parameters}, with more details regarding the convergence and parameter selection included in Appendix~\ref{appendix:convergence}.

\begin{table}[pos=htp]
\caption{Hyperparameters used in the experiments for the GMM and KNN algorithms. R-GMM refers to robust GMM.}
\begin{tabular}{lll}
Algorithm & Hyperparameter                        & Values              \\\hline
GMM       & K                                     & Estimated using BIC \\
GMM       & Regularization parameter (scree test) & $10^{-5}$           \\
R-GMM     & $\text{th}$                                  & 0.5                 \\
R-GMM     & $\alpha$                              & 40                  \\
KNN       &Number of neighbors k                & 5                  
\end{tabular}
\label{table:parameters}
\end{table}

\subsection{Data imputation}\label{sec:data_imputation}

This section evaluates the imputation performance of the proposed GMM method when applied to time series of vegetation indices for crop monitoring. In particular, we test the robustness of the proposed approach to the amount of missing data, and in a second step to the amount of outlier by introducing samples coming from different crop types.

\subsubsection{Imputation results obtained by varying the amount of S2 images affected by missing values}\label{sec:results_rapeseed}

The dataset used in this study is relatively exempt of errors coming from the parcel data (e.g, less than 1\% of errors in the crop type reported) or the features (e.g., few undetected clouds) as detailed in \citet{Mouret_2021}. As a consequence, this dataset is a good start to test the imputation methods in controlled conditions. 

\begin{figure}[htp]
    \centering
    \includegraphics[width=1\textwidth]{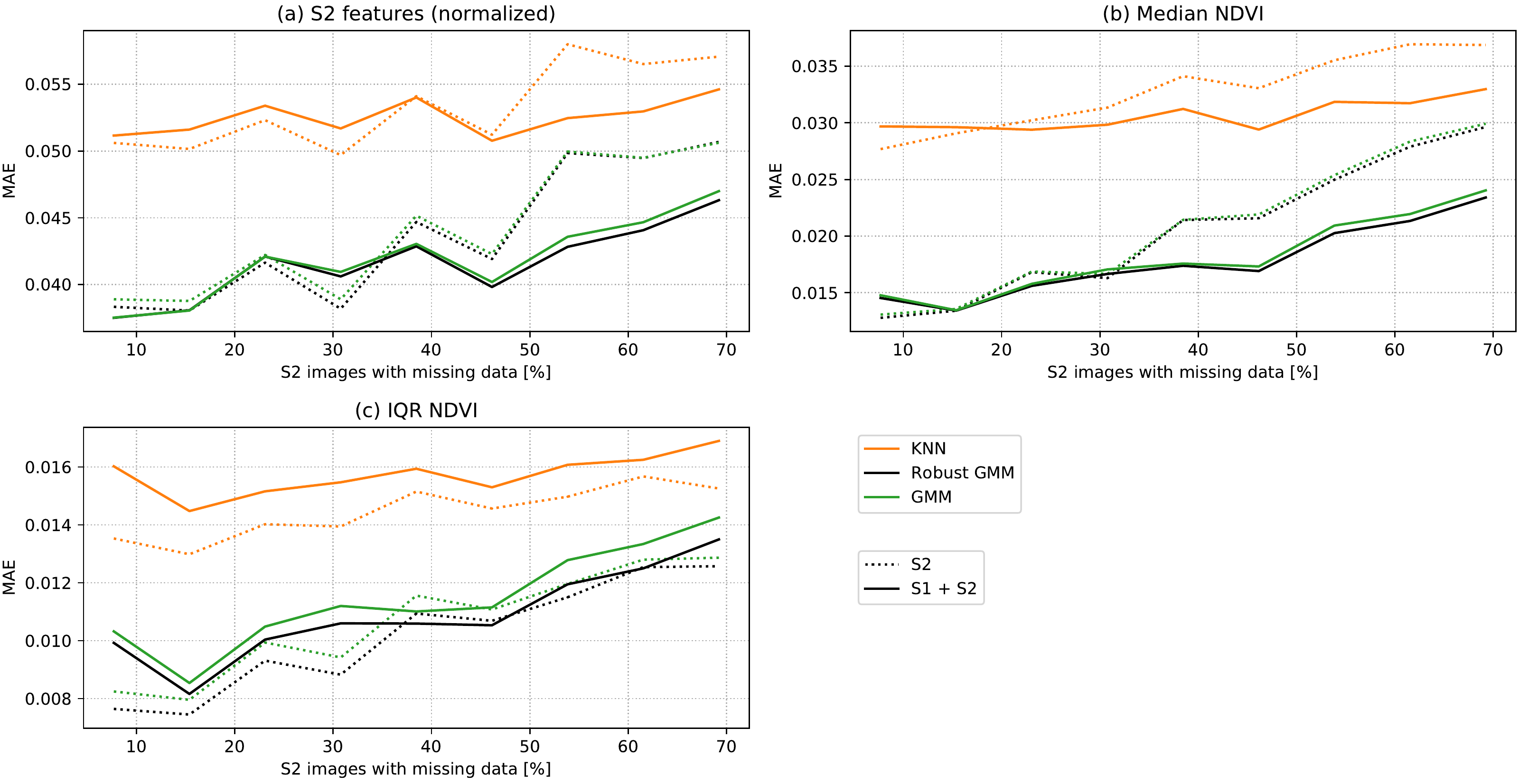}
    \caption{MAE for rapeseed vegetation indices versus the percentage of missing images. X-axis: percentage of S2 images with missing values. Y-axis: MAE for (a) the normalized S2 features (all the S2 indicators are considered), (b) the median of NDVI and (c) the IQR of the NDVI (computed at the parcel level). Results in dotted lines are obtained using S2 features only whereas solid lines correspond to the joint use of S1 and S2 data. The results are averaged after 50 MC runs.}
    \label{fig:MAE_vs_percentage_missing_rapeseed}
\end{figure}

The influence of the amount of missing data on the imputation results was tested by varying the percentage of S2 images affected by missing values, as depicted in \autoref{fig:MAE_vs_percentage_missing_rapeseed}. All the results were obtained by averaging the outputs of $50$ MC runs. We recall that for each S2 image with missing data, 50\% of the parcels were randomly chosen in the database and their corresponding features were removed. The MAE obtained for all the S2 features is depicted in \autoref{fig:MAE_vs_percentage_missing_rapeseed}(a) whereas \autoref{fig:MAE_vs_percentage_missing_rapeseed}(b) and (c) show specifically the MAE of the median and IQR NDVI. Note that in \autoref{fig:MAE_vs_percentage_missing_rapeseed}(a), the S2 features are scaled in the range [0,1] to be able to have comparable results (\textit{e.g.}, MCARI/OSAVI features are not normalized), which can lead to MAE greater than those obtained in natural scale. One can observe that the GMM imputation algorithm outperforms the KNN imputation with good reconstructions even with a high amount of missing data. Results obtained with the classical GMM are close to those obtained with the robust GMM in these experiments. Looking specifically at the median NDVI (\autoref{fig:MAE_vs_percentage_missing_rapeseed}(b)), it appears that using S1 data is particularly useful, especially when there is a high amount of missing S2 features (the same observation was made for the median statistics of the other VIs studied). Indeed, when the percentage of S2 images with missing data is higher than 40\%, one can observe that the MAE is significantly lower when using S1 images (plain curves) to reconstruct median NDVI features (this is also true for S2 features in general, as highlighted in \autoref{fig:MAE_vs_percentage_missing_rapeseed}(a)). For example, when the percentage of missing S2 images is equal to 70\%, the difference in MAE obtained when imputing median NDVI values with and without S1 images is around 0.006 when using GMM imputation methods (corresponding to a reduction of the MAE close to 20\%). The reconstruction of IQR statistics is however not favored by the use of S1 data, as shown in \autoref{fig:MAE_vs_percentage_missing_rapeseed}(c). For this statistics, the robust GMM provides a lower MAE than the classical GMM.

\subsubsection{Imputation results obtained by introducing samples coming from different crop types}\label{sec:contamination_experiment}

In practice, errors or noise can contaminate the feature matrix with samples that are very different from the rest of the data. In that case, GMM learning can be more difficult and lead to inaccurate imputations. To investigate the sensitivity of the imputation method to the presence of irrelevant samples, agricultural parcels with a different crop type than rapeseed were included into the rapeseed dataset (these crop types mainly correspond to wheat, maize and barley). The features of these parcels were extracted using field boundaries coming from the French Land Parcel Identification System (LPIS) \citep{BARBOTTIN2018281}, which is available in open license. 

\begin{figure}[htp]
    \centering
    \includegraphics[width=1\textwidth]{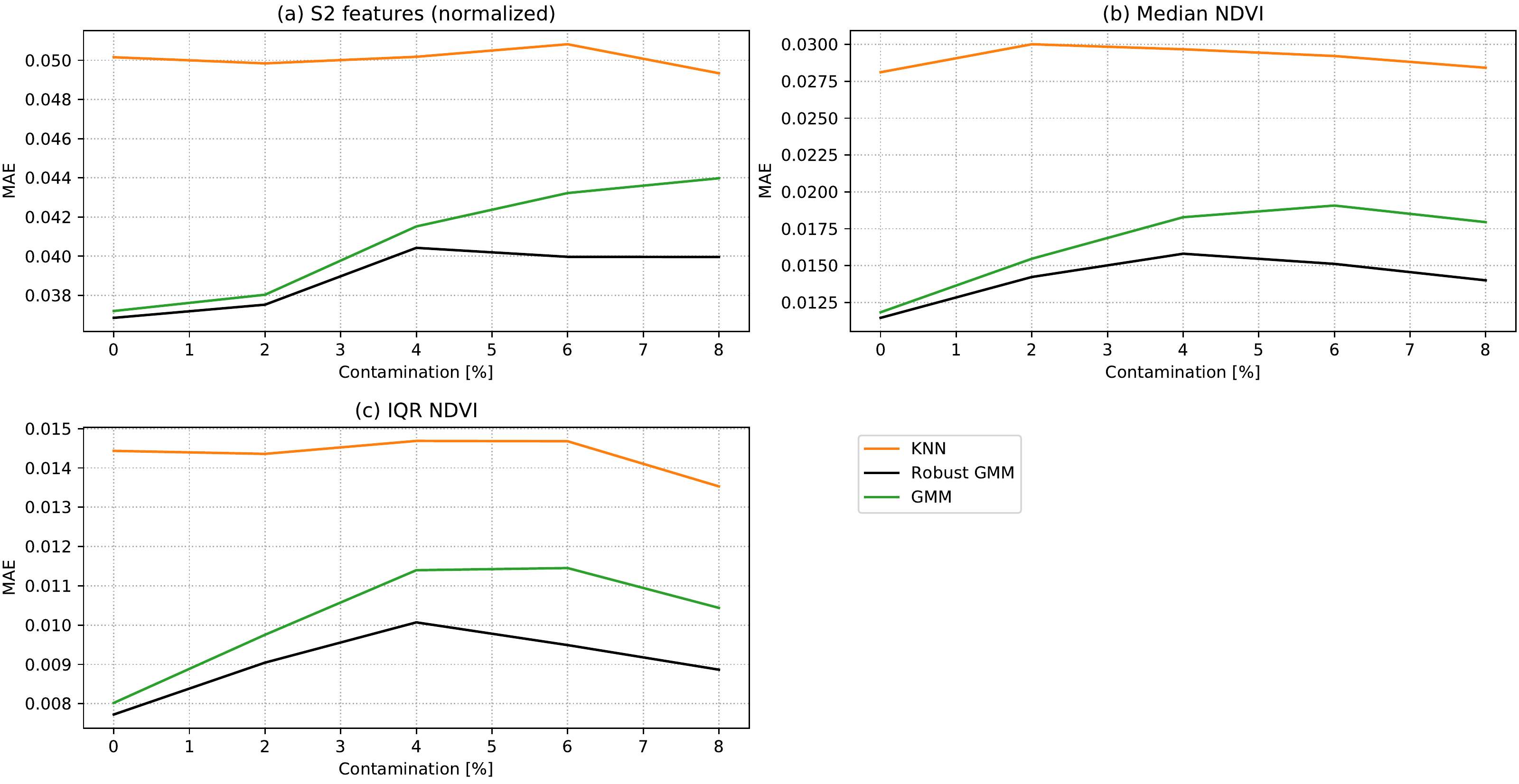}
    \caption{Median of MAE versus the percentage of contamination in the dataset (i.e., coming from non-rapeseed crops) after 50 MC runs for (a) the normalized S2 features (all the S2 indicators are considered), (b) the median of NDVI and (c) the IQR of NDVI (computed at the parcel level). Results are obtained using S1 and S2 features jointly. For each MC run, the percentage of missing data has been fixed: three S2 images (23\%) have missing data affecting 50\% of the parcels.}   \label{fig:mae_contamination}
\end{figure}

Imputation results obtained on the rapeseed parcels by varying the percentage of contamination (i.e., the percentage of non-rapeseed parcels in the dataset) are provided in \autoref{fig:mae_contamination}, showing the median of the MAE computed using $50$ MC runs. The median of the MAE is used here since some extreme MAE values are obtained when using the standard GMM imputation due the presence of non-rapeseed parcels (contrary to the robust GMM). For each run, there are three random S2 images with missing values affecting 50\% of the parcels (note that the MAE is computed using the rapeseed parcels only). Using the robust GMM imputation is particularly useful in that case, with an MAE almost stable with respect to the percentage of irrelevant samples in the dataset (it is especially true for the NDVI statistics). Note that the standard GMM imputation is highly impacted by the presence of outliers in the dataset and can lead to large errors, with reconstruction sometimes worse than those obtained using the KNN approach. Consequently, using the robust GMM imputation is recommended in practice, especially if the dataset contains some irrelevant samples.

\subsection{Application to the detection of anomalous crop development in the presence of missing data}\label{sec:result_application}

\begin{figure}[htp]
    \subfloat[]{\includegraphics[width=0.39\textwidth]{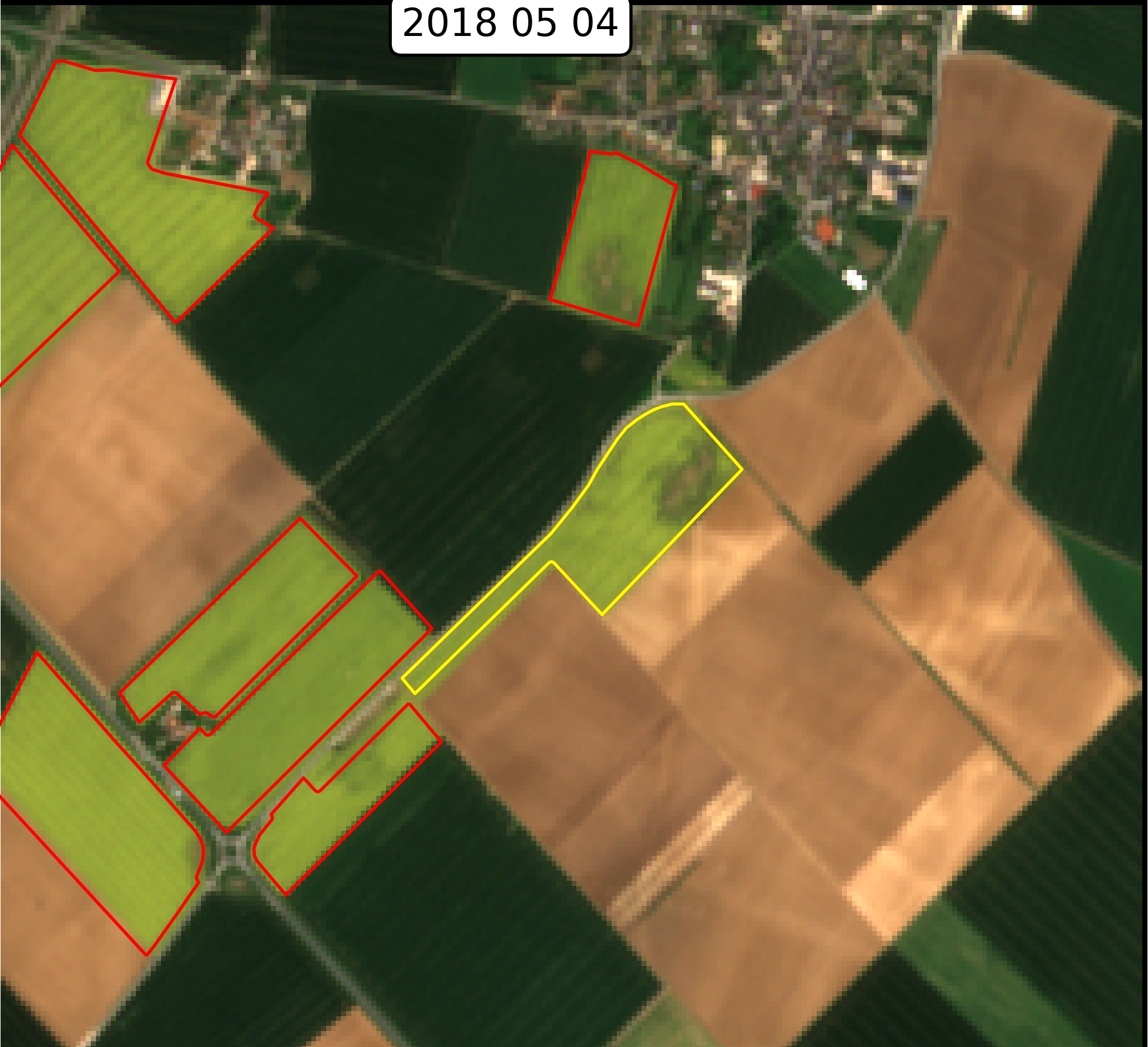}}
    \subfloat[]{\includegraphics[width=0.5\textwidth]{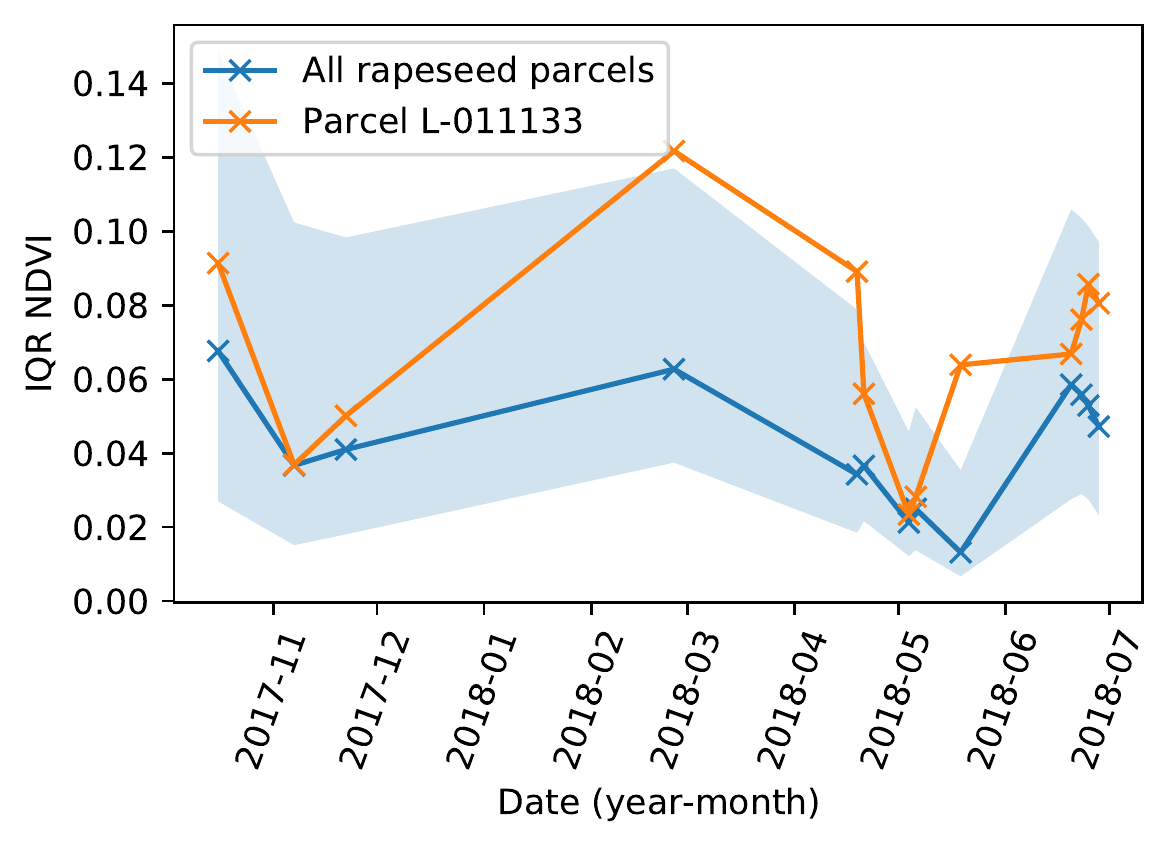}}
    \caption{(a) A rapeseed parcel (yellow boundaries) affected by heterogeneity, the image was acquired in May $2018$. (b) Interquartile Range (IQR) of the NDVI time series for the yellow parcel (orange line). The blue line is the median value of the whole dataset. The blue area is filled between the 10th and 90th percentiles. The orange line clearly shows an abnormal behaviour of the parcel due here to heterogeneity problems.}
    \label{fig:ex_het}
\end{figure}

Detecting potential anomalies in the development of crop parcels is an important problem in crop monitoring. Such detection can be useful for farmers or agricultural cooperatives to automatically detect parcels with potential problems, without a need for on-site visits. Typical crop anomalies can be grouped into 4 categories: growth anomalies, heterogeneity problems, database problems (\textit{i.e.}, wrong crop type or inaccurate boundaries reported in the database) and false positives (\textit{e.g.}, due to undetected clouds or shadows). An example of heterogeneous parcel is depicted in \autoref{fig:ex_het} (a further analysis showed that a part of the parcel was damaged during winter). Each parcel was labeled by an agronomic expert as true positive (relevant anomaly to be detected) or false positive (not relevant for crop monitoring).

\begin{figure}[htp]
    \centering
    \includegraphics[width=0.5\textwidth]{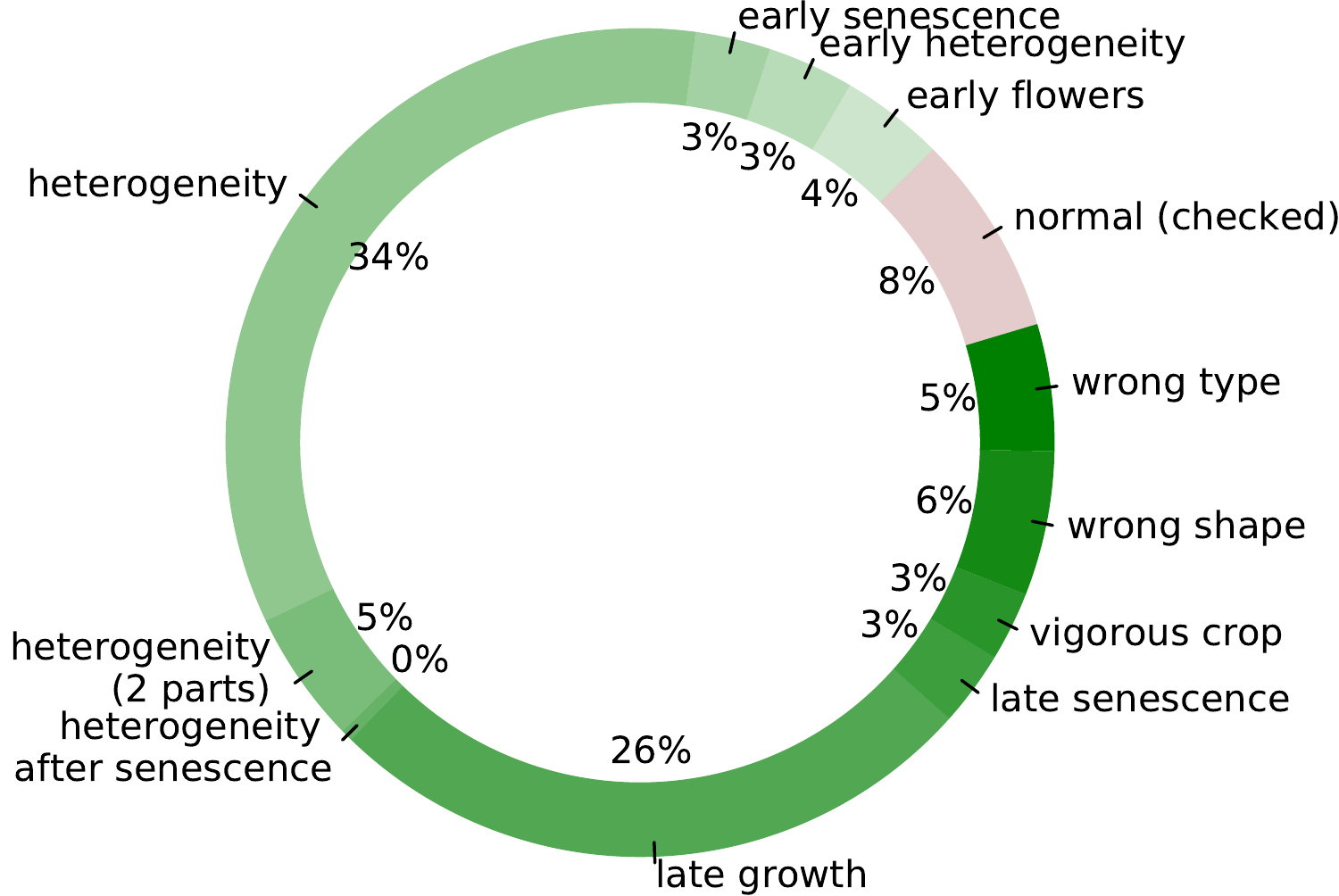}
    \caption{Distribution of the detected rapeseed parcels in each outlier category when using the IF algorithm with an outlier ratio of $20\%$.}
    \label{fig:pie_IF_02}
\end{figure}

\newpage
Following the method proposed in \citet{Mouret_2021}, parcels with abnormal behavior are detected using the Isolation Forest algorithm, which computes anomaly scores using the feature matrix (whose construction is detailed in Section~\ref{sec:features}). The strongest anomalies are generally related to errors in the crop type (as simulated in Section~\ref{sec:contamination_experiment}), whereas other anomalies are parcels with abnormal phenological development. While it was shown that this method is useful to detect relevant anomalies in the crop development, one main problem is that this method cannot be applied in the presence of missing data. As a consequence, we propose to use an imputation method before the outlier detection step, in order to consider images partially covered by clouds. The distribution of the rapeseed parcels detected as abnormal using the IF algorithm with an outlier ratio of $20\%$ is displayed in \autoref{fig:pie_IF_02}.

\begin{figure}[htp]
    \centering
    \includegraphics[width=1\textwidth]{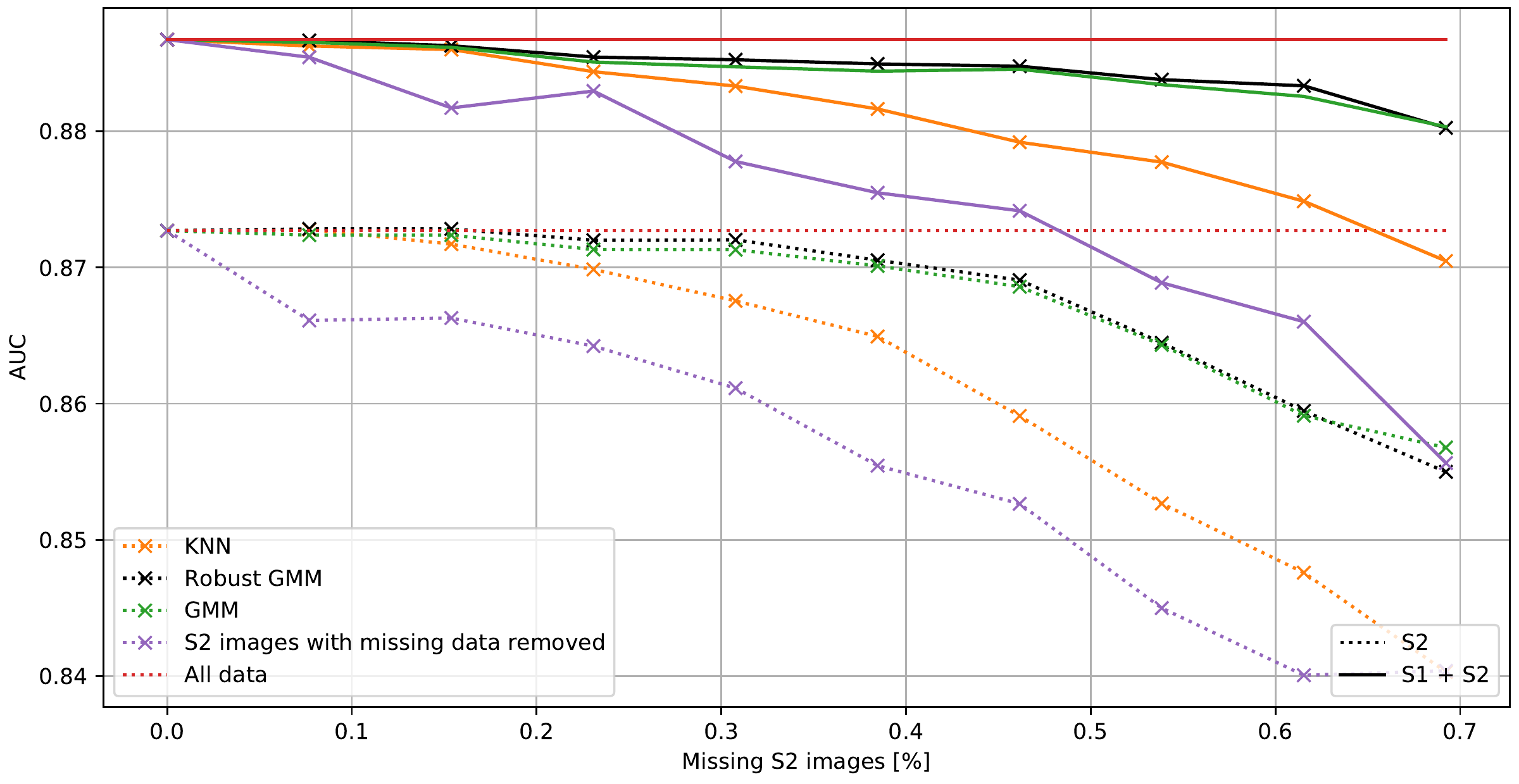}
    \caption{Area under the precision vs. outlier ratio curve (AUC) w.r.t. the percentage of cloudy S2 images (50\% of the parcels in a cloudy S2 image have missing data, i.e., do not contain S2 features). Results in dotted lines are obtained using S2 features only whereas solid lines correspond to the joint use of S1 and S2 data. All results are averaged using 50 MC runs.}
    \label{fig:AUC_rapeseed}
\end{figure}

This section evaluates the influence of missing values on the detection results resulting from the application of the IF algorithm. The AUC values (the higher the better) obtained by varying the amount of S2 images affected by missing data are displayed in \autoref{fig:AUC_rapeseed} (more details on this metric were provided in Section~\ref{sec:metrics}). It can be observed that accurate results are obtained with AUC greater than $0.84$, even with a high percentage of missing data in the dataset. In particular, the best results are obtained using S1 and S2 data jointly and a reconstruction with the GMM (both robust and non-robust versions perform similarly in that case). It is interesting to note that discarding S2 images affected by missing values yields a reduced detection performance, since some of the anomalies cannot be detected anymore.  Thus, the imputation methods are able to reconstruct the VI with sufficient accuracy to detect the abnormal crop parcels and important information on the parcel behavior seems to be lost without the reconstruction step.

\section{Discussion}\label{sec:discussion}

This section provides some comments about the results obtained in this paper. Additional experiments (available in the supplementary material) are also discussed.

\subsection{Analysis of the presented results}

The experiments conducted in this study show that 1) GMM imputations outperform the KNN method 
2) using a robust GMM is of crucial importance in the presence of strong outliers. Hence, our results confirm the interest of using outlier detection techniques as standard preprocessing steps in remote sensing, as also recommended for instance in \citep{Pelletier2017} for the classification of land cover. Note that the obtained results are coherent with the literature: an MAE of 0.0281 was obtained in \citep{Yu_2021} for the reconstruction of NDVI in crop vegetation, an MAE of 0.038 was obtained in \citep{Garioud_2020} for the reconstruction of NDVI for grassland parcels and MAE varying from $0.035$ to $0.042$ (depending on the region analyzed) was obtained in \citep{Garioud_2021} for agricultural parcels. While these results provide quantitative values for comparison purposes, important differences have to be highlighted: existing studies generally focus on NDVI time series acquired at the pixel-level and do not analyze crops at the parcel level, as proposed in this paper. Moreover, some of these studies focus on the regression of SAR time series towards NDVI \citep{Garioud_2021}. Nevertheless, obtaining an MAE close to $0.013$ (resp. to $0.019$) when imputing the median NDVI for rapeseed (resp. wheat) crops is encouraging (see Tables S1 and S2 in the supplementary materials).

The interest of using a combination of S1 and S2 features was confirmed by our experiments. In particular, using S1 features is interesting to reconstruct more accurately S2 features and thus ensures a better detection of crop anomalies. Two specific examples illustrating the interest of using S1 data are provided in the supplementary material for rapeseed and wheat parcels (Figures S2 and S3). The heterogeneity of a parcel (summarized using IQR at the parcel-level) is less linked to S1 data, which confirms previous results given in \citep{Mouret_2021}. It was also observed that using various features extracted from S2 data helps to reconstruct missing data in NDVI time series when compared to using NDVI only. The interest of using S1 and S2 images jointly has also been confirmed for the detection of anomalous crop development in the presence of missing data, supporting the previous results found in \cite{Mouret_2021}.

When removing features from one of the S2 image (Figures S4 and S5 in the supplementary material), it appears that some specific stages of the growing season are more difficult to reconstruct, with differences observed for rapeseed and wheat crops. For rapeseed crops, the first S2 acquisition (October 10) is challenging to reconstruct. One explanation is that at this date some fields are not sowed yet whereas others are already vigorous, leading to a higher dispersion of the parcel indicators. The higher MAE obtained for S2 data acquired in February can be explained as follows: 1) S2 images before and after this date correspond to very distant dates 2) crop parcels can be more or less affected by winter, which again leads to a larger dispersion of the indicators. Regarding wheat crops, the high reconstruction errors observed for the data acquired in June 18 can be explained by the beginning of the senescence, which leads to abrupt changes in the crop behavior. 

\subsection{Other imputation methods}

Other strategies for the imputation of missing data were tested without bringing any improvement compared to the proposed method (see Figure S6 in the supplementary material). Some observations are briefly provided below.

Gap filling methods (linear interpolation, spline interpolation and Whittaker smoother) \citep{Cai_2017} perform overall poorly compared to the methods investigated in this paper, which is mainly due to the sparsity of S2 acquisitions, confirming the results found in \citep{Yu_2021}. Moreover, when applied to the detection of abnormal crop development, smoothing methods tend to decrease the accuracy of the detection results. Other benchmark imputation methods were tested, such as Multiple Imputation by Chained Equations (MICE) proposed in \citet{Buuren2011} and implemented in the Python library scikit-learn \citep{scikit-learn}. Similarly to the KNN imputation, this method provides reconstruction results significantly less accurate than those obtained using the proposed GMM imputation algorithm. Deep learning methods were also tested without success (see an experimental example with Generative Adversarial Networks \citep{Yoon_GAIN_2018} in Figure S7 of the supplementary materials). The poor results obtained with these methods (which include LSTM, denoising autoencoders \citep{Vincent_2008} or Generative Adversarial Networks \citep{Yoon_GAIN_2018}) are probably due to the small number of samples in the dataset, which confirms the observations made in \cite{Yoon_GAIN_2018} (Figure 2).

Finally, we considered some outlier detection methods that do not need to impute missing data. It is the case with the IF algorithm, which can be extended to handle missing values without imputation using the strategies studied in \citep{Zemicheal2019, cortes2019imputing}. This type of strategy is appealing since it drastically reduces the computation time when compared to GMM-based methods. However we observed that these methods are sensitive to the amount of missing values in the dataset and can lead to poor results. Moreover, having access to reliable reconstructed time series is interesting for crop monitoring since it allows the user to analyze with more details the behavior of an abnormal parcel. 

\subsection{Regularization techniques for GMM}

GMM are subject to the curse of dimensionality \citep{BOUVEYRON_review}. This problem was confirmed in our application, especially due to the small number of parcels compared to the high number of features. The regularization of \citet{BOUVEYRON_article} used in this paper provided the best results overall. 

Another classical regularization consists of adding a sparsity constraint to the precision matrices, which can be solved using the graphical lasso algorithm \cite{friedman_sparse_2008}, which has been adapted to the missing data problem \citep{ruan_2011}. However, using such regularization yielded poor results for the reconstruction of vegetation indices. The sparsity of the precision matrix is due to conditionally independent variables, which is not the case in the proposed feature vector gathering the same features acquired at different time instants. The sparsity of the covariance matrices was also investigated using the method proposed in \citep{fop_2019} without improving the results obtained with the $[a_{ij}bQ_{i}d_{i}]$ model suggested in \citet{BOUVEYRON_article}.

\newpage
\section{Conclusion}\label{sec:conclusion}

This paper studied an imputation method based on Gaussian Mixture Models (GMM) for the reconstruction of remote sensing time series constructed from vegetation indices (VI) associated with Sentinel-2 (S2) data. One contribution of this paper is to propose a method able to reconstruct simultaneously various time series, coming from different VI whose statistics have been computed at the parcel-level. These statistics (here, the median and interquartile range) are well suited for crop monitoring since they can characterize efficiently the parcel behaviors, \textit{e.g.}, to detect abnormal growth or heterogeneity problems. This paper showed that using a GMM imputation to reconstruct missing values in the feature matrix performs significantly better than other reference methods such as the k-nearest neighbors or the Multiple Imputation by Chained Equations (MICE)).

Another contribution of this paper is to propose a robust GMM imputation method, which attributes weights to each sample based on the outlier scores resulting from the Isolation Forest algorithm. Samples with high outlier scores have reduced weights, limiting their impact on the estimation of the GMM parameters. Using the proposed robust GMM method instead of the standard GMM imputation method is particularly useful in the presence of irrelevant samples contaminating the dataset. For operational services, we then recommend to use this robust version since it consistently provides reconstruction results similar or better than the standard GMM imputation method.

The experiments conducted in this paper have shown that using Sentinel-1 (S1) features in addition to S2 features improves GMM imputation results in different cases. In particular, using S1 images can be interesting when the amount of missing S2 images is important (above 40\%) and for some specific S2 features such as the median NDVI of the parcels. This indicator can be reconstructed with good accuracy (mean absolute error (MAE) close to $0.013$ for rapeseed crops and to $0.020$ for wheat crops), even with a high amount of missing data (\textit{e.g.}, for rapeseed parcels, the MAE is close to $0.020$ even when 70\% of the S2 images have 50\% of the parcels affected by missing data). 

An application to the detection of anomalous crop development in presence of missing data was also investigated. Using S1 and S2 images jointly provided best results for this application. Using a Gaussian mixture model (GMM) for the reconstruction of missing data provided detection results significantly better than with KNN imputation. Note that discarding images affected by clouds leads to poor results for this application.

Further investigations will be conducted to determine whether other regularizations could be applied to GMM to improve the imputation results, for instance by finding an adapted structure for the covariance matrices or by reducing the dimensionality of the dataset. Moreover, since GMM are good models for vegetation indices, other applications, such as forecasting, clustering or classification, would deserve to be investigated. In particular, the automatic classification of the different anomalies could be considered as in \citet{leon2022anomaly}. Adding external information such as climate data could also be relevant to reconstruct more efficiently the various VI, since interesting results were obtained for the reconstruction of NDVI time series \citep{VUOLO2017202, Yu_2021}. Moreover, adding other type of image data such as unmanned aerial vehicles (UAV) imagery could be also interesting to improve the quality of the imputations \citep{TANG2021105895, HUANG2021102590}.

\printcredits

\begin{appendices}
\setcounter{equation}{0}%
\renewcommand{\theequation}{\thesection.\arabic{equation}}

\setcounter{figure}{0}%
\renewcommand{\thefigure}{\thesection.\arabic{figure}}

\section{The EM algorithm for GMM}

\subsection{The standard EM algorithm (without missing data)}\label{appendix:standard_EM}

Given a feature matrix $\boldsymbol{X}$, we assume that each row of this matrix is distributed according to a mixture of $K$ Gaussian distributions. The corresponding log-likelihood can be expressed as (up to an additive constant):
\begin{equation}\label{eq:LL}
    \log \mathcal{L}(\boldsymbol{\theta}; \boldsymbol{X}) = \sum_{n=1}^{N} \log \left( \sum_{k=1}^{K}\pi_k \mathcal{N}(\boldsymbol{x}_n | \boldsymbol{\mu}_k, \boldsymbol{\Sigma}_k) \right),
\end{equation}
\noindent where $N$ is the number of samples in the dataset, $\boldsymbol{x}_n$ is a specific sample, $\mathcal{N}(\boldsymbol{x}_n | \boldsymbol{\mu}_k, \boldsymbol{\Sigma}_k)$ is the probability density function (PDF) of the multivariate normal distribution and $\boldsymbol{\theta} =$ $ \{\pi_1, ..., \pi_k, $ $\boldsymbol{\mu}_1, ..., \boldsymbol{\mu}_k, \boldsymbol{\Sigma}_1, ..., \boldsymbol{\Sigma}_K \}$ contains the parameters to be estimated. These parameters are the mean vectors $\boldsymbol{\mu}_k$, the covariance matrices $\boldsymbol{\Sigma}_k$ and the mixing coefficients $\pi_k$ ($0 < \pi_k < 1, \sum_{k=1}^{K}\pi_k = 1$) of the GMM. The maximization of~\eqref{eq:LL} with respect to (w.r.t.) $\boldsymbol{\theta}$ being complex, it is classical to introduce labels $\boldsymbol{z}= \{\boldsymbol{z}_1, ..., \boldsymbol{z}_n \}$ indicating the Gaussians associated with the observed vectors $\boldsymbol{x}_1,...,\boldsymbol{x}_n$ (the maximization of the likelihood is straightforward for known labels) and the complete log-likelihood:
\begin{equation}
    \log \mathcal{L}_c(\boldsymbol{\theta}; \boldsymbol{X}, \boldsymbol{z}) = \sum_{n=1}^{N} \sum_{k=1}^{K} z_{nk} \log \left[ \pi_k \mathcal{N}(\boldsymbol{x}_n | \boldsymbol{\mu}_k, \boldsymbol{\Sigma}_k) \right],
\end{equation}
\noindent where $z_{nk}= 1$ if the vector $\boldsymbol{x}_n$ belongs to the $k$th component of the GMM and $z_{nk}= 0$ otherwise. After an appropriate initialization of $\boldsymbol{\theta}$, the EM algorithm aims at maximizing the conditional expectation of $\log \mathcal{L}_c(\boldsymbol{\theta}; \boldsymbol{X}, \boldsymbol{z})$ in an iterative fashion until convergence. The expectation step (E-step) computes the expectation of the complete log-likelihood conditionally to the current set of the mixture parameters, $\boldsymbol{\theta}^{(t)}$:
\begin{equation}
    E[\log \mathcal{L}_c(\boldsymbol{\theta}; \boldsymbol{X}, \boldsymbol{z})) | \boldsymbol{\theta}^{(t)}] = \sum_{n=1}^{N} \sum_{k=1}^{K} \gamma_{nk} \log \left[ \pi_k \mathcal{N}(\boldsymbol{x}_n | \boldsymbol{\mu}_k, \boldsymbol{\Sigma}_k) \right],
\end{equation}
\noindent where $\gamma_{nk} = E[z_n = k | \boldsymbol{x}_n, \boldsymbol{\theta}^{(t)}]$ is referred to as responsibilities. The maximization step (M-step) maximizes $E[\log \mathcal{L}_c(\boldsymbol{\theta}; z) | \boldsymbol{\theta}^{(t)}]$ w.r.t. $\boldsymbol{\theta}$ to provide an updated parameter vector $\boldsymbol{\theta}^{(t+1)}$:
\begin{equation}
    \boldsymbol{\theta}^{(t+1)} = \textrm{arg max}_{\boldsymbol{\theta}} E[\log \mathcal{L}_c(\boldsymbol{\theta}; z) | \boldsymbol{\theta}^{(t)}].
\end{equation}
\noindent For brevity, we will denote $\boldsymbol{\theta}^{(t)} = \boldsymbol{\theta}$, \textit{i.e.}, $\boldsymbol{\mu}_k^{(t)} = \boldsymbol{\mu}_k$, $\boldsymbol{\Sigma}_k^{(t)} = \boldsymbol{\Sigma}_k$ and $\pi_k^{(t)} = \pi_k$ the current set of parameters in what follows.

\textit{E-step:} in practice the E-step reduces to the computation of the responsibilities $\gamma_{nk}$, which are also the probabilities that the sample $\boldsymbol{x}_n$ has been generated by the $k$th Gaussian component:
\begin{equation}
    \gamma_{nk} = \frac{\pi_k \mathcal{N}(\boldsymbol{x}_n | \boldsymbol{\mu}_k, \boldsymbol{\Sigma}_k)}{\sum_{j=1}^K \pi_j \mathcal{N}(\boldsymbol{x}_n | \boldsymbol{\mu}_j, \boldsymbol{\Sigma}_j)}.
\end{equation}

\textit{M-step:} the parameters are re-estimated using the updated responsibilities:
\begin{equation}
    \boldsymbol{\mu}_k^{(t+1)} = \frac{1}{N_k}\sum_{n=1}^{N} \gamma_{nk} \boldsymbol{x}_n, \; \boldsymbol{\Sigma}_k^{(t+1)} = \frac{1}{N_k}\sum_{n=1}^{N}\gamma_{nk}(\boldsymbol{x}_n - \boldsymbol{\mu}_k)(\boldsymbol{x}_n - \boldsymbol{\mu}_k)^T, \; \pi_k^{(t+1)} = \frac{N_k}{N},
\end{equation}
\noindent with $N_k = \sum_{n=1}^{N} \gamma_{nk}$. The EM algorithm is stopped when the log-likelihood or the parameter values does not change significantly or after a fixed number of iterations.

\subsection{Regularization of the covariance matrices}\label{sec:regularization}
Learning the parameters of a GMM can be subject to instabilities, especially when the covariance matrices are ill-conditioned (in some extreme cases, the covariance matrix cannot even be inverted). A heuristic strategy for regularizing a covariance matrix consists in adding a small constant to its diagonal elements during the estimation (\textit{e.g.}, this regularization  is proposed in the Python library scikit learn \cite{scikit-learn}). Alternatively, \citet{BOUVEYRON_article} studied different regularization techniques adapted to the estimation of covariance matrices for high dimensional problems. In this study, we have considered the model referred to as $[a_{ij}bQ_{i}d_{i}]$  (see \citet{BOUVEYRON_article} for details). The idea behind this model is to use an eigendecomposition of the covariance matrices $\boldsymbol{\Sigma}_k = Q_k \Delta_k Q_k^T$ and set the smallest eigenvalues to the same constant $b_k=b$ (which can be justified when the data are obtained in a common acquisition process). This operation significantly reduces the number of model parameters to estimate, which is valuable to fight against the curse of dimensionality. The scree test is used to find the number of eigenvalues to be set to the constant value $b$ (see \citet{BOUVEYRON_article} for details).

\newpage
\subsection{Proposed algorithm}\label{appendix:algo}

\begin{algorithm}[htp!]
\caption{Proposed robust EM algorithm for GMM in the the incomplete data case.}\label{alg:em_proposed}
\textbf{Input:} Data  $\{\textbf{\textit{x}}_n\}_{n=1}^N$, $K$ the number of clusters \\
\textbf{Output:} Clustering labels $\mathcal{Z}=\{z_n\}_{n=1}^N$, parameters $\boldsymbol{\theta} = \{\pi_1, ..., \pi_k, $ $\boldsymbol{\mu}_1, ..., \boldsymbol{\mu}_k, \boldsymbol{\Sigma}_1, ..., \boldsymbol{\Sigma}_K \}$ and imputed samples $\hat{\boldsymbol{x}}$
\begin{algorithmic}[1]
\State For each sample, identify observed and missing components $o$ and $m$;
\State Initialize $\boldsymbol{\theta}^{(0)}$; (\textit{e.g.}, randomly or using the kmeans algorithm)
\State $t \leftarrow 1$

\While{not convergence}

\For{$1 \leq k \leq K$:}\Comment{\textbf{E-step}}

    \Comment{Compute observed responsibilities:}
    \State $\gamma_{nk}^{(t)} = \frac{\pi_k^{(t-1)} \mathcal{N}(\boldsymbol{x}_n^o, \boldsymbol{\mu}_k^{o(t-1)}, \boldsymbol{\Sigma}_k^{oo(t-1)})}{\sum_{j=1}^{K} \pi_j^{(t-1)} \mathcal{N}(\boldsymbol{x}_n^o, \boldsymbol{\mu}_j^{o(t-1)}, \boldsymbol{\Sigma}_j^{oo(t-1)})} $
    \medskip
    
    \Comment{Compute conditional expectations:}
    \State $\hat{\boldsymbol{\mu}}_{nk}^{m(t)} = \boldsymbol{\mu}_k^{m(t-1)} + \boldsymbol{\Sigma}_k^{mo(t-1)}(\boldsymbol{\Sigma}_k^{oo(t-1)})^{-1}(\boldsymbol{x}_n^o - \boldsymbol{\mu}_k^{o(t-1)})$
    \medskip
    \State $\hat{\boldsymbol{\Sigma}}_{nk}^{mm(t)} = \boldsymbol{\Sigma}_k^{mm(t-1)} - \boldsymbol{\Sigma}_k^{mo(t-1)}(\boldsymbol{\Sigma}_k^{oo(t-1)})^{-1}\boldsymbol{\Sigma}_k^{mo(t-1)}$

    \State Fill in: $\hat{\boldsymbol{x}}_{nk}^{(t)} \leftarrow [\boldsymbol{x}_i^{o}, \hat{\boldsymbol{\mu}}_{nk}^{m(t)}]$ and  $\hat{\boldsymbol{\Sigma}}_{nk}^{(t)} \leftarrow \begin{pmatrix} 
	\boldsymbol{0}^{oo} & \boldsymbol{0}^{om} \\
	\boldsymbol{0}^{mo} & \hat{\boldsymbol{\Sigma}}_{nk}^{mm(t)}
	\end{pmatrix}$
\EndFor

\For{$1 \leq k \leq K$:}\Comment{\textbf{M-step}}
\medskip
\State $w_{n} = \frac{1}{1 + \exp{[\alpha(\textrm{score}_{\textrm{IF}}(\boldsymbol{\hat{\boldsymbol{x}}}_n^{(t)}) - \text{th})}]}$\Comment{Compute outlier weights using the IF algorithm}
\medskip

\State $\pi_{k}^{(t)} = \frac{1}{N} \sum_{n=1}^N \gamma_{nk}^{(t)}$
\medskip
\State $\boldsymbol{\mu}_{k}^{(t)} = \frac{\sum_{n=1}^{N} w_n \gamma_{nk}^{(t)} \hat{\boldsymbol{x}}_{nk}^{(t)}}{\sum_{n=1}^{N} w_n \gamma_{nk}^{(t)}}$
\medskip
\State $\boldsymbol{\Sigma}_k^{(t)} = \frac{\sum_{i=1}^{N} w_n^2 \gamma_{nk}^{(t)}\left[(\hat{\boldsymbol{x}}_n^{(t)} - \hat{\boldsymbol{\mu}}_k^{(t)})(\hat{\boldsymbol{x}}_n^{(t)} - \hat{\boldsymbol{\mu}}_k^{(t)})^T + \hat{\boldsymbol{\Sigma}}_{nk}^{(t)} \right]}{\sum_{n=1}^{N} w_n^2 \gamma_{nk}^{(t)}}$
\medskip

\EndFor

\State $t \leftarrow t+1 $

\EndWhile
\State Set $z_i$ as the index $k$ that has the maximimum $p_{ik}$.
\State \textbf{Final imputation:} $\hat{\boldsymbol{x}}_n^m = \sum_{k=1}^K \gamma_{nk}^{(t)} \hat{\boldsymbol{x}}_{nk}^{(t)}$
\end{algorithmic}
\end{algorithm}

\section{Parameter tuning and convergence}\label{appendix:convergence}

\subsection{GMM imputation algorithm}

The number of Gaussians $K$ in the GMM was estimated using the Bayesian Information Criterion (BIC) as suggested in \citet{BOUVEYRON_review}. This estimation provided good results without a need for manually choosing the number of components, which can be difficult in practice (especially for an unsupervised task). For the regularization of the covariance matrix, the stopping criterion of the scree test was set to $10^{-5}$ (see \cite{BOUVEYRON_article} for more details on the scree test). We observed that a too small value (typically lower than $10^{-6}$) can lead to unstable results whereas too high values (typically $10^{-3}$) lead to a deterioration of the imputation results.

The parameters of the weighting function $w_n$ are the threshold $\text{th}$ and the slope $\alpha$. The threshold was fixed to $\text{th}=0.5$, which is  a natural value to separate outliers and inliers when using the IF algorithm \citet{Liu2012}. The slope parameter was fixed to $\alpha=40$ by cross validation. Small changes in these parameters did not have a significant impact on the imputation results.

\begin{figure}[pos=htp]
    \centering
    \includegraphics[width=0.5\textwidth]{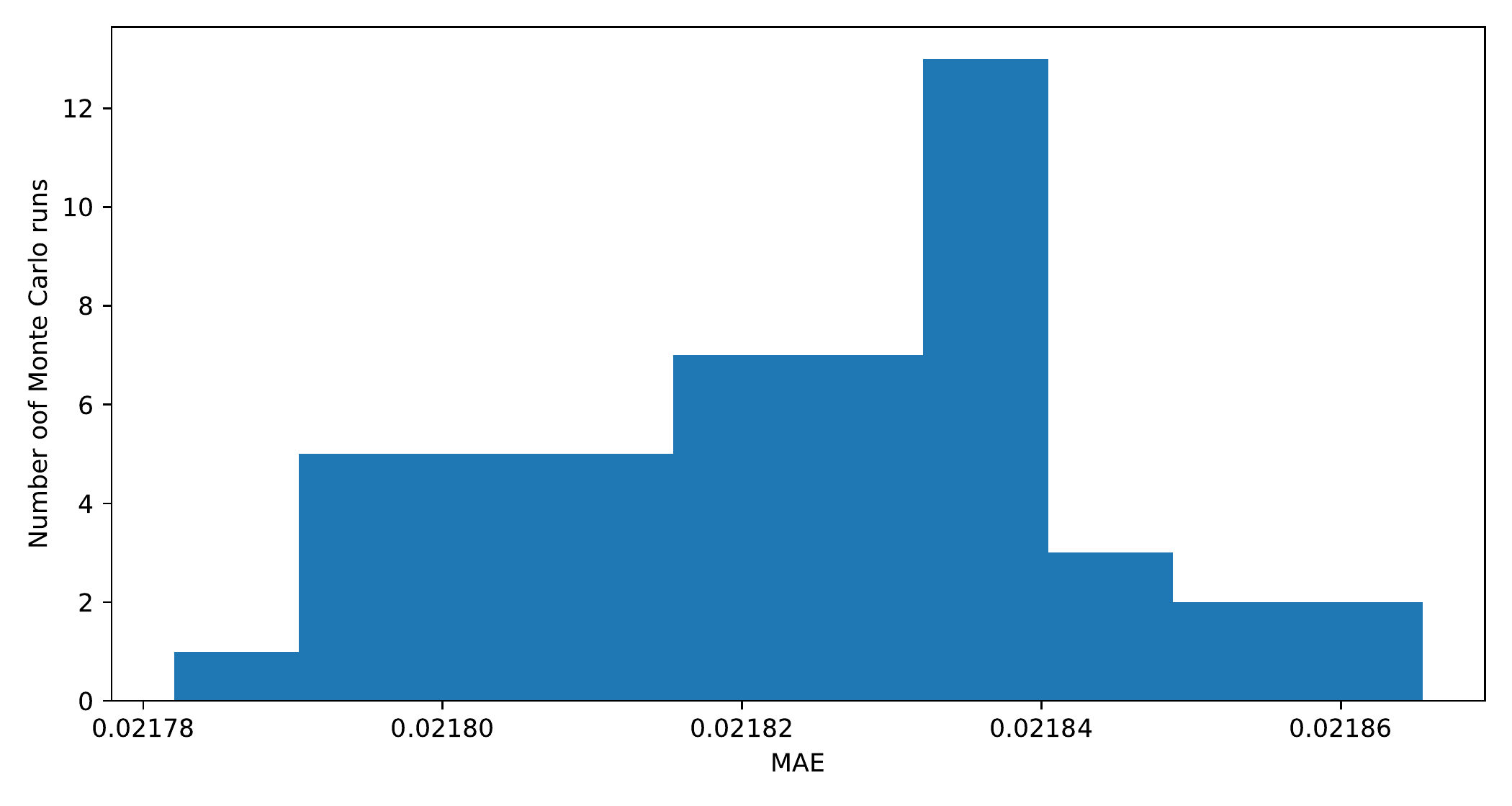}
    \caption{Histogram of MAE obtained after 50 Monte Carlo runs (with different initializations) on the same dataset.}
    \label{fig:std_imputation}
\end{figure}

The outputs of the EM algorithm depend on its initialization, which is detailed in what follows. The EM algorithm was initialized by the output of the K-means algorithm with $K$ centroids chosen equally likely in the dataset. This initialization yields a fast convergence of the EM algorithm obtained in less than $10$ iterations. The EM algorithm was stopped when the difference between two consecutive values of the log-likelihood was less than $10^{-3}$. In order to analyze the sensitivity of the algorithm to its initialization, we ran $50$ Monte Carlo (MC) simulations of the EM algorithm using the same dataset (1 S2 image covered by clouds, 50\% of the parcels affected by missing values) with different random initializations and imputed the missing values. The distribution of the MAE obtained for these Monte Carlo runs (evaluated using all the reconstructed features for the parcels with missing data) is displayed in \autoref{fig:std_imputation}, showing that the values of MAE are very similar, varying in the interval [0.02178, 0.02186]. These results indicate that the EM algorithm is not very sensitive to its initialization for the reconstruction of VI at the parcel level.

\subsection{KNN imputation algorithm}

The KNN imputation method available in the Python library Scikit-Learn \citep{scikit-learn} (version 0.24) (named ``KNNimputer'') was used as a benchmark. The number of  nearest neighbors was fixed to $k=5$. Changing the value of this parameter in a neighborhood did not have a huge effect on the reconstruction results. The contribution of each neighbor was weighted by the inverse of its distance to the sample to be imputed, similarly to the configuration used in \citep{Albughdadi2017}.

\end{appendices}

\bibliographystyle{cas-model2-names}

\bibliography{references}

\end{document}


This section provides complementary experiments conducted on rapeseed and wheat crops regarding the reconstruction of vegetation indices computed using Sentinel-2 images. 

\section{Details regarding the wheat dataset}

\begin{figure}[ht]
\centering
\includegraphics[width=0.5\textwidth]{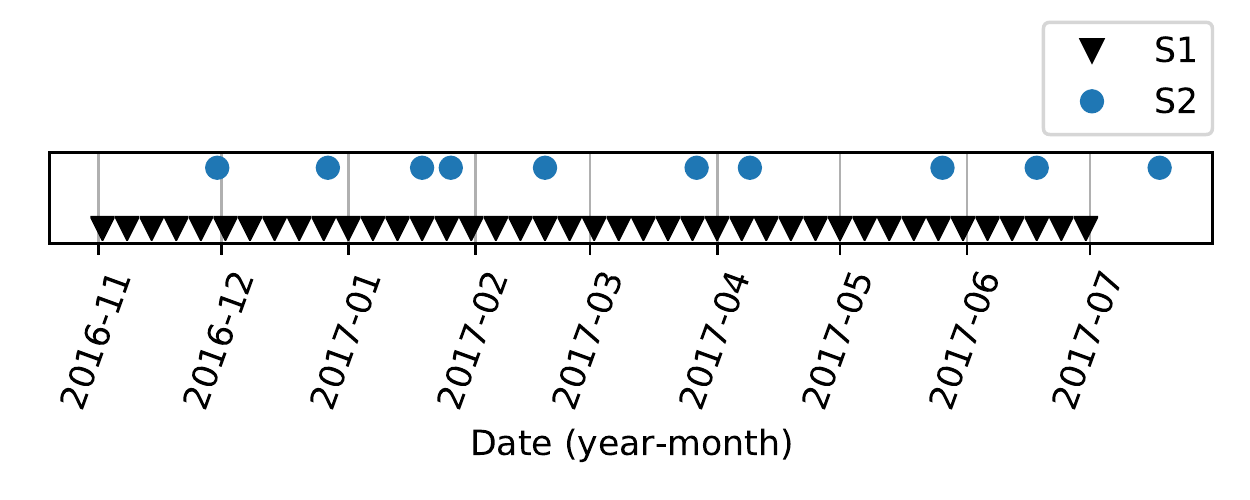}
\caption{Each marker corresponds to the acquisition date of a used image for the growing season 2016/2017.}
\label{fig:calendar}
\end{figure}

In addition to the rapeseed dataset presented in the main document, $3361$ wheat parcels were analyzed during the growing season 2016/2017. Overall, $10$ S2 and $41$ S1 images were selected for this growing season, with acquisition dates displayed in \autoref{fig:calendar}.

\section{Examples of data imputation for rapeseed and wheat parcels}

\subsection{Rapeseed parcel}
\begin{figure}[htp!]
    \centering
    \includegraphics[width=1\textwidth]{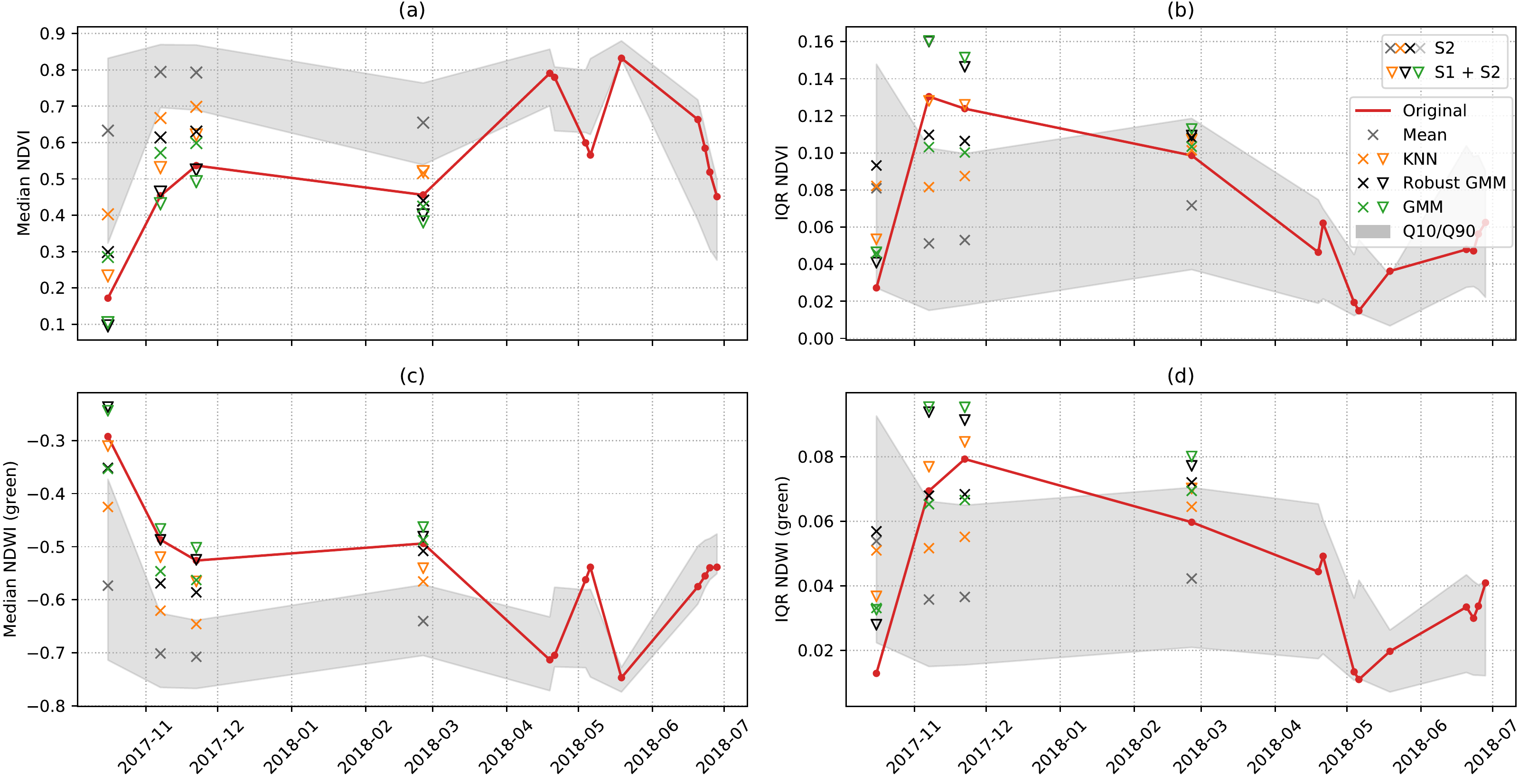}
    \caption{For a specific rapeseed parcel, imputation of (a) median NDVI, (b) IQR NDVI, (c) median NDWI (green) and (d) IQR NDWI (green). The crosses correspond to imputations obtained by using S2 images only whereas triangles correspond to the joint use of S1 and S2 features. The gray area is filled between the 10th and 90th percentiles values of the whole dataset. 50\% of the parcels are affected by missing values.}
    \label{fig:example_rapeseed}
\end{figure}

To have an easier appreciation of the challenges related to the data imputation in S2 data, examples of reconstruction of 4 different time series (median NDVI (a), IQR NDVI (b), median NDWI (c) and IQR NDWI (d)) for a specific parcel are displayed in \autoref{fig:example_rapeseed}. For that experiment, the first 4 S2 acquisitions have missing values and 50\% of the parcels are affected, making the imputation problem particularly difficult. It can be observed that the analyzed parcel (red curve) has a late development during the first part of the growing season, which is an atypical behavior making the reconstruction task even harder. Here, the mean imputation is displayed (gray crosses) to show that this method can be problematic if the values to be imputed are unusual. Overall, the robust GMM (black) is generally more accurate, with almost perfect reconstruction for the median features in November. The reconstruction of the IQR values tends to be overestimated in that case, even if the heterogeneous behavior is captured. In this extreme example, the interest of using S1 data (triangles) is particularly visible: since no data is available during the first part of the growing season, using this information is of crucial importance for all the tested methods. Finally, one can observe that the KNN imputation can provide good results (especially when using S1 data), but is generally less accurate (\textit{e.g.}, median NDVI is constantly underestimated).

\newpage
\subsection{Wheat parcel}

\begin{figure}[htp]
    \centering
    \includegraphics[width=1\textwidth]{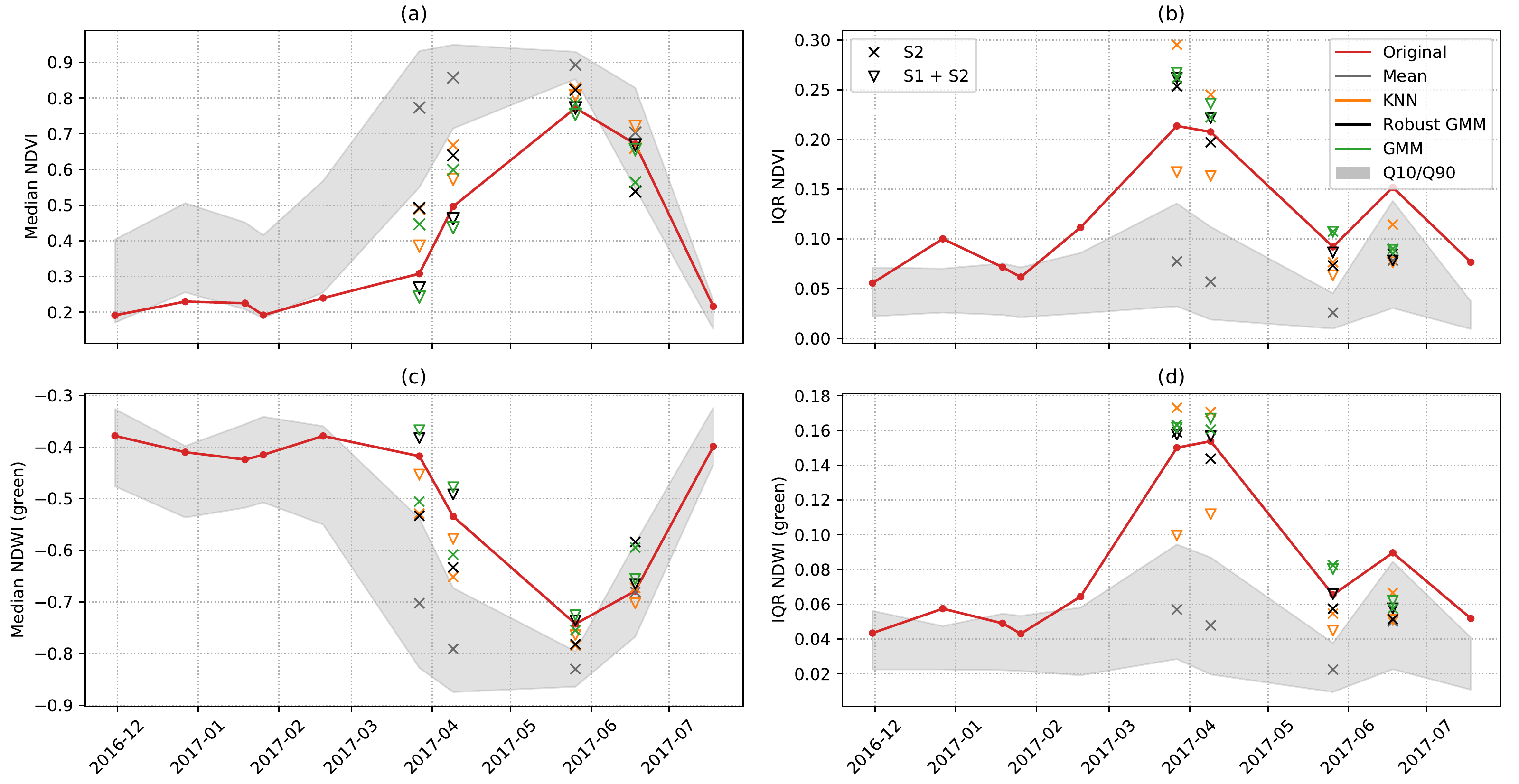}
    \caption{For a specific rapeseed parcel, imputation of (a) median NDVI, (b) IQR NDVI, (c) median NDWI (green) and (d) IQR NDWI (green). The crosses correspond to imputations obtained by using S2 images only whereas triangles correspond to the joint use of S1 and S2 features. The gray area is filled between the 10th and 90th percentiles values of the whole dataset. 50\% of the parcels are affected by missing values.}
    \label{fig:example_wheat}
\end{figure}

\autoref{fig:example_wheat} provides an example similar to the one displayed in \autoref{fig:example_rapeseed} but for a wheat parcel with a late and heterogeneous development. One can notice that the robust GMM imputations are close to the original values and that again using S1 data is helpful to capture the low vigor of the crop. 

\newpage
\section{Day by day imputation}

\subsection{Rapeseed crops}
\begin{figure}[htp!]
    \centering
    \includegraphics[width=0.6\textwidth]{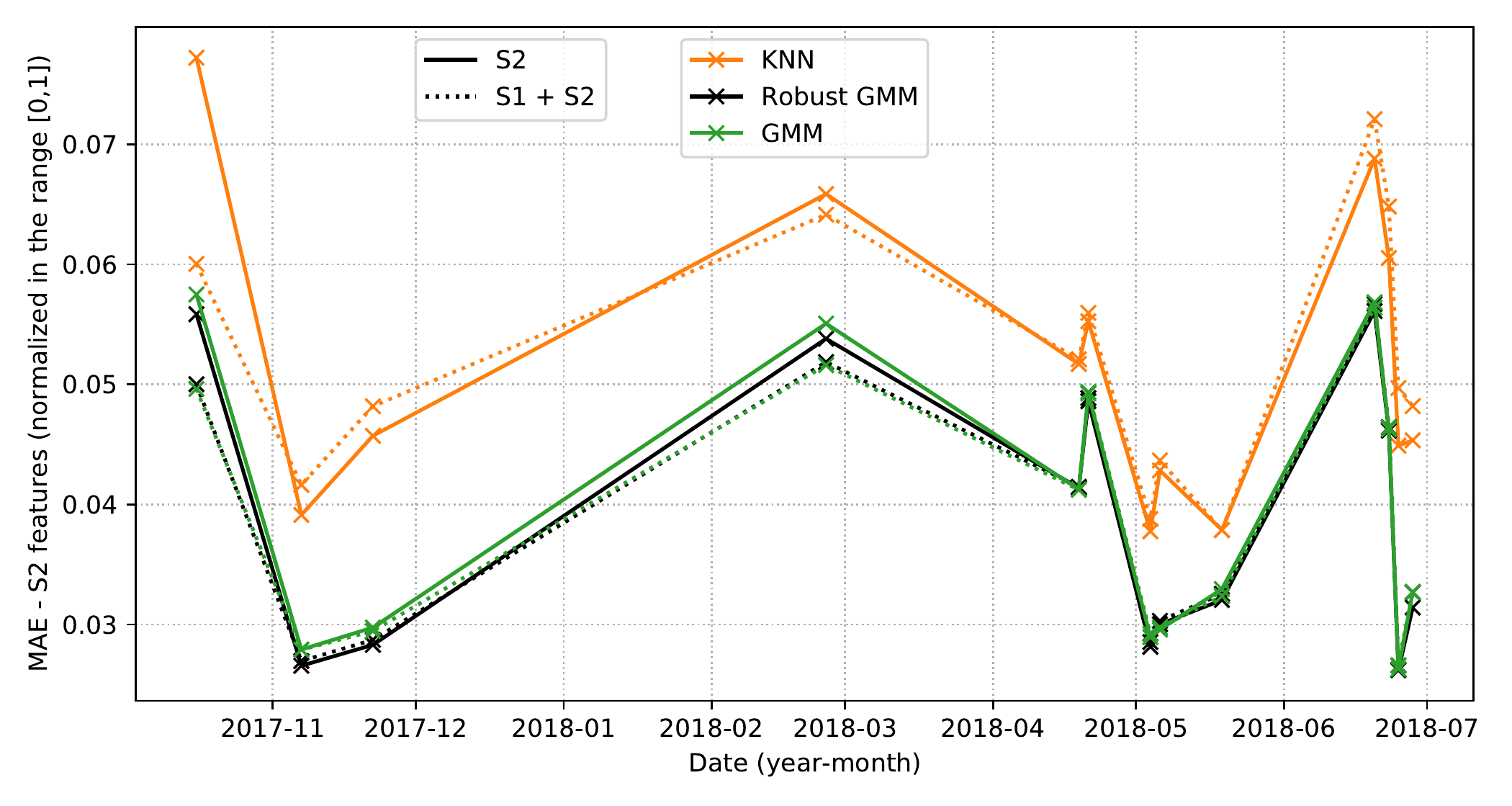}
    \caption{Rapeseed crops are analyzed. X-axis: S2 acquisition with missing values: for each acquisition with missing data, 50\% of the parcels are affected. Y-axis: MAE of the normalized S2 features the parcels. The solid lines results are obtained using only the S2 data, whereas dashed lines were obtained using both the S1 and S2 data. Results are averaged after 50 iterations}
    \label{fig:MAE_rapeseed_day_by_day}
\end{figure}

An experiment was conducted by removing S2 features at each S2 acquisition (50\% of the crop parcels are affected).
The results obtained for the normalized S2 features are depicted in \autoref{fig:MAE_rapeseed_day_by_day} (similar results have been observed when looking at specific features and are not plotted here for conciseness). In this example, two dates are more difficult to impute: the first acquisition (during sowing) and the data acquired at the end of February during winter. On the other hand, some stages of the growing season can be reconstructed with significantly lower MAE. Again, GMM imputation outperforms KNN imputation. Moreover, using S1 data is also useful to improve the results (particularly for the most difficult imputations).

\newpage
\subsection{Wheat crops}

\begin{figure}[htp]
    \centering
    \includegraphics[width=0.6\textwidth]{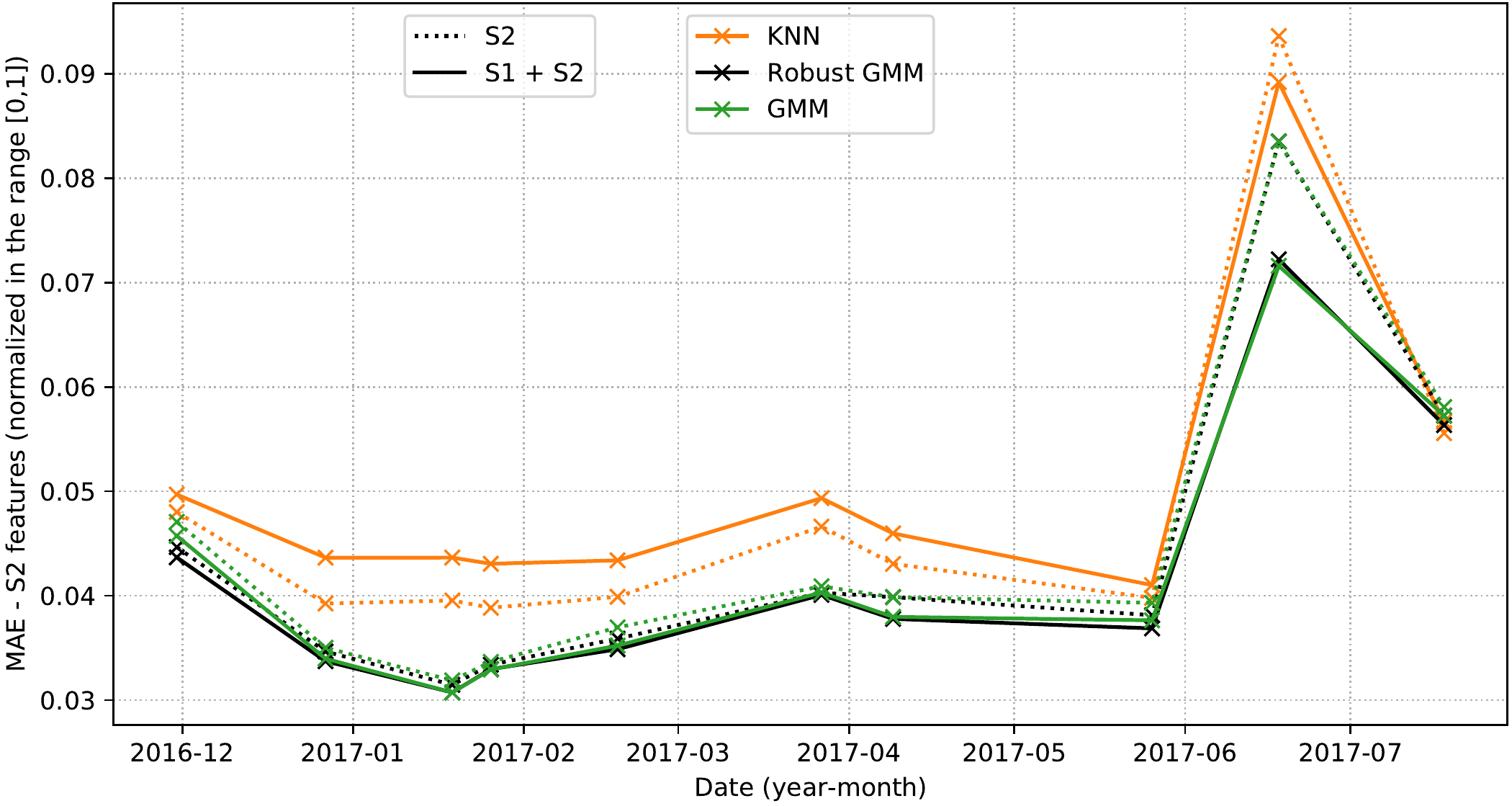}
    \caption{Analysis of wheat crops. X-axis: S2 acquisitions with missing values: for each acquisition with missing data, 50\% of the parcels are affected. Y-axis: MAE for normalized S2 features  (all the S2 indicators are considered). The solid lines are obtained using S2 data only, whereas the dashed lines were obtained using both S1 and S2 data. Results are averaged using 50 Monte Carlo runs, each time 50\% of the parcels are affected by missing data.}
    \label{fig:MAE_wheat_day_by_day}
\end{figure}

The experiments presented in \autoref{fig:MAE_rapeseed_day_by_day} were also conducted for wheat crops. Results are displayed in \autoref{fig:MAE_wheat_day_by_day} for the normalized S2 signal. The robust GMM always provides the best reconstructions, in particular for the most difficult imputations. The features extracted from the S2 images acquired in June are the most difficult to reconstruct. For this acquisition, using S1 data improves the imputation significantly. The other experiments conducted on rapeseed crops lead to the same conclusions for the wheat parcels (not shown here). 

\section{Detailed results for a specific S2 acquisition with missing data}

The MAE, RMSE and coefficient of determination ($R^2$) were computed by averaging the results obtained after 500 simulations. For each simulation, all the S1 and S2 data were used and 50\% of the parcels have missing values at a same random S2 acquisition. The obtained scores for rapeseed crops are provided in \autoref{table:result_rapeseed}. These results confirm that the robust GMM provides the best reconstructions overall (it is also the case when compared to the classical GMM). Some features are harder to reconstruct: it is particularly the case for the MCARI/OSAVI statistics, and more generally for the IQR of the different VI. IQR is not well captured by the S1 data and has less smooth time variations, which could explain this results. 

\begin{table}[H]
\caption{Regression scores (MAE, RMSE, $R^2$) obtained on the rapeseed dataset (200 MC simulations). Each time, 50\% of the parcels have missing values at one S2 acquisition. S1 and S2 features are used to impute missing values. Standard deviation (std) is added in parenthesis. KNN and Robust-GMM (R-GMM) are compared, best results are in bold.}
\centering
\begin{tabular}{lllllll}
                        & \multicolumn{2}{c}{MAE (std)} & \multicolumn{2}{c}{RMSE (std)} & \multicolumn{2}{c}{$R^2$ (std)} \\ \hline
Feature / Algorithm     & KNN          & R-GMM          & KNN           & R-GMM          & KNN                   & R-GMM                  \\
\hline
median(NDVI)            &        0.029 (0.008)     &      \textbf{0.013 (0.007)}          &      0.042 (0.009)         &         \textbf{0.021 (0.009)}       &         0.71 (0.17)              &        \textbf{0.92 (0.09)}                \\
median($\textnormal{NDWI}_{\textnormal{GREEN}}$) &      0.021 (0.007)        &       \textbf{0.010 (0.005)}        &       0.030 (0.007)        &       \textbf{0.016 (0.006) }        &             0.64 (0.15)          &           \textbf{0.88 (0.10) }            \\
median($\textnormal{NDWI}_{\textnormal{SWIR}}$)  &     0.029 (0.008)         &        \textbf{0.012 (0.007)}        &      0.043 (0.007)         &          \textbf{0.019 (0.009)}      &        0.77 (0.14)               &            \textbf{0.95 (0.04)}            \\
median(GRVI)            &      0.027 (0.007)        &           \textbf{0.014 (0.007)}     &    0.037 (0.008)           &       \textbf{0.020 (0.009)}         &            0.72 (0.16)           &     \textbf{0.89 (0.10)}                   \\
median(MCARI/OSAVI)     &        133 (55)      &        \textbf{79 (46)}        &       180 (66)        &       \textbf{117 (64)}         &      0.61 (0.23)                 &           \textbf{0.81 (0.16)  }           \\ \hline
IQR(NDVI)               &      0.015 (0.005)        &       \textbf{0.008 (0.005)}         &       0.024 (0.008)        &        \textbf{0.014 (0.007)}       &      0.58 (0.10)                 &        \textbf{0.83 (0.13) }               \\
IQR($\textnormal{NDWI}_{\textnormal{GREEN}}$)    &      0.009 (0.003)        &       \textbf{0.006 (0.003)}         &      0.016 (0.004)         &     \textbf{0.010 (0.005)}           &            0.52 (0.08)           &          \textbf{0.80 (0.14) }             \\
IQR($\textnormal{NDWI}_{\textnormal{SWIR}}$)     &       0.018 (0.003)       &        \textbf{0.009 (0.005)}        &       0.028 (0.005)        &      \textbf{0.014 (0.007)}          &           0.57 (0.12)            &        \textbf{0.86 (0.14)}                \\
IQR(GRVI)               &      0.013 (0.004)        &           \textbf{0.008 (0.004)}     &    0.019 (0.005)           &       \textbf{0.13 (0.006)}         &         0.51 (0.15)              &  \textbf{0.76 (0.18)}                      \\
IQR(MCARI/OSAVI)        &       62 (22)       &         \textbf{42 (20)}       &        93 (30)       &        \textbf{65 (30)}        &   0.40 (0.18)                    &    \textbf{0.68 (0.21)}                   \\\hline
\end{tabular}
\label{table:result_rapeseed}
\end{table}

The same experiment was done for wheat crops whose results are reported in \autoref{table:result_wheat}.

\begin{table}[H]
\caption{Regression scores (MAE, RMSE, $R^2$) obtained on the wheat dataset (200 MC simulations). Each time, 50\% of the parcels have missing values at one S2 acquisition. S1 and S2 features are used to impute missing values. Standard deviation (std) is added in parenthesis. KNN and Robust-GMM (R-GMM) are compared, best results are in bold.}
\centering
\begin{tabular}{lllllll}
                        & \multicolumn{2}{c}{MAE (std)} & \multicolumn{2}{c}{RMSE (std)} & \multicolumn{2}{c}{$R^2$ (std)} \\ \hline
Feature / Algorithm     & KNN          & R-GMM          & KNN           & R-GMM          & KNN                   & R-GMM                  \\
\hline
median(NDVI)            &        0.032 (0.013)     &      \textbf{0.020 (0.010)}          &      0.044 (0.015)         &         \textbf{0.029 (0.013)}       &         0.70 (0.27)              &        \textbf{0.81 (0.26)}                \\
median($\textnormal{NDWI}_{\textnormal{GREEN}}$) &      0.026 (0.008)        &       \textbf{0.018 (0.008)}        &       0.035 (0.009)        &       \textbf{0.024 (0.006) }        &             0.67 (0.30)          &           \textbf{0.79 (0.29) }            \\
median($\textnormal{NDWI}_{\textnormal{SWIR}}$)  &     0.031 (0.012)         &        \textbf{0.020 (0.009)}        &      0.042 (0.015)         &          \textbf{0.029 (0.012)}      &        0.69 (0.21)               &            \textbf{0.82 (0.21)}            \\
median(GRVI)            &      0.032 (0.020)        &           \textbf{0.024 (0.018)}     &    0.042 (0.027)           &       \textbf{0.033 (0.024)}         &            0.63 (0.18)           &     \textbf{0.75 (0.27)}                   \\
median(MCARI/OSAVI)     &        127 (93)      &        \textbf{106 (113)}        &       206 (215)        &       \textbf{178 (228)}         &      0.51 (0.25)                 &           \textbf{0.64 (0.28)  }           \\ \hline
IQR(NDVI)               &      0.015 (0.008)        &       \textbf{0.010 (0.007)}         &       0.024 (0.009)        &        \textbf{0.017 (0.01)}       &      0.41 (0.18)                 &        \textbf{0.68 (0.26) }               \\
IQR($\textnormal{NDWI}_{\textnormal{GREEN}}$)    &      0.011 (0.004)        &       \textbf{0.008 (0.004)}         &      0.018 (0.006)         &     \textbf{0.012 (0.005)}           &            0.32 (0.19)           &          \textbf{0.59 (0.27) }             \\
IQR($\textnormal{NDWI}_{\textnormal{SWIR}}$)     &       0.015 (0.007)       &        \textbf{0.010 (0.005)}        &       0.022 (0.008)        &      \textbf{0.016 (0.008)}          &           0.29 (0.14)            &        \textbf{0.60 (0.23)}                \\
IQR(GRVI)               &      0.014 (0.011)        &           \textbf{0.012 (0.010)}     &    0.021 (0.014)           &       \textbf{0.18 (0.013)}         &         0.35 (0.20)              &  \textbf{0.55 (0.27)}                      \\
IQR(MCARI/OSAVI)        &       \textbf{83 (82)}       &         88 (117)       &        163 (210)       &        \textbf{154 (225)}        &   0.29 (0.20)                    &    \textbf{0.44 (0.28)}                   \\\hline
\end{tabular}
\label{table:result_wheat}
\end{table}

\section{Imputation results using other algorithms}

\begin{figure}[htp]
    \centering
    \includegraphics[width=1\textwidth]{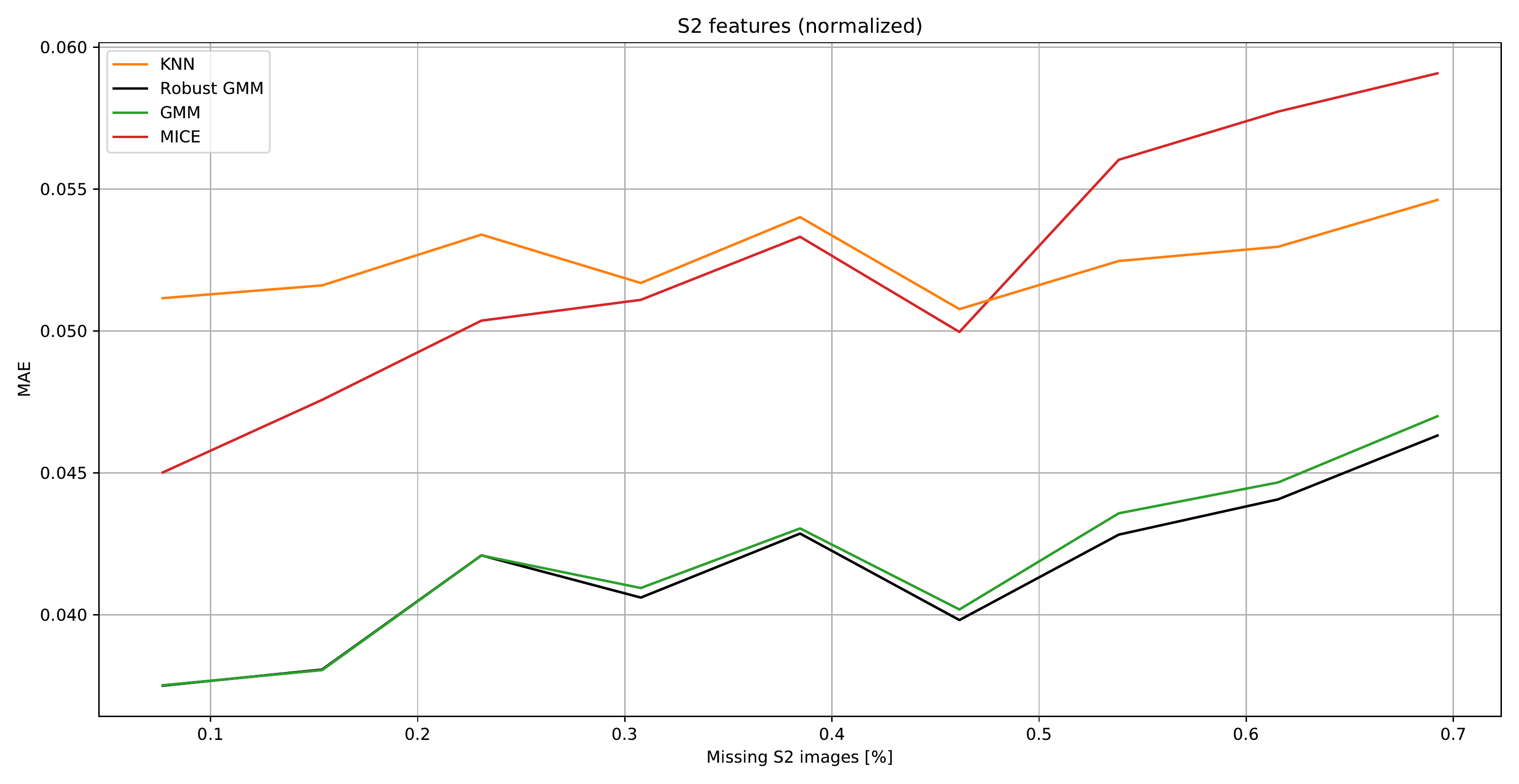}
    \caption{Analysis of rapeseed crops. X-axis: percentage of S2 with missing values. Y-axis: MAE for the normalized S2  features (all the S2 indicators are considered). Results are obtained using S1 and S2 data jointly. The results are averaged using $50$ MC runs, each time 50\% of the parcels are affected by missing data for each S2 image with missing data.}
    \label{fig:MICE}
\end{figure}

The experiment presented in Figure 6 of the main document was conducted using various imputation algorithms. Results obtained after adding the Multiple Imputation by Chained Equations (MICE) algorithm are displayed in \autoref{fig:MICE} (using S1 and S2 data jointly). Overall, the MICE algorithm provides imputation slightly better than the KNN algorithms (except when the amount of missing S2 images is greater than 50\%). In any cases, using GMM imputation is significantly better.

\begin{figure}[htp!]
    \centering
    \subfloat[]{\includegraphics[width=1\textwidth]{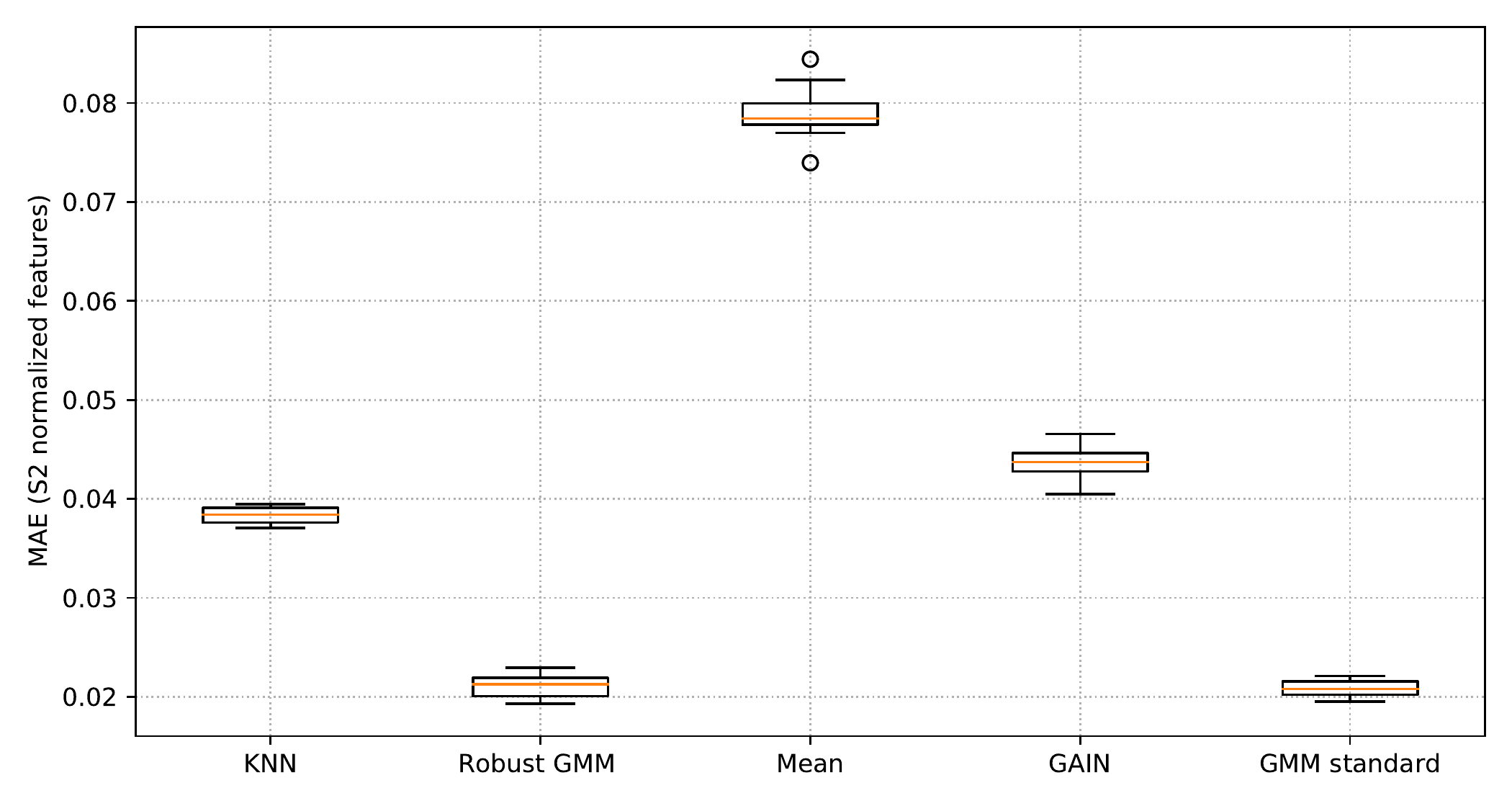}}
    \caption{MAE obtained on S2 features (normalized). For each of the 50 MC simulations, one S2 image has 20\% of parcels with missing data.}
    \label{fig:GAIN}
\end{figure}

Similarly, experiments were conducted to compare our approach to the Generative Adversarial Imputation Networks (GAIN) framework \citep{Yoon_GAIN_2018}. For each MC run, one S2 image (randomly chosen) has 20\% of its parcels affected by missing data (50 MCs experiments were run). The results are reported in \autoref{fig:GAIN}. It can be observed that the GAIN framework is not competitive in that configuration. One obvious explanation to these results is the limited number of samples in our dataset, which is in accordance with observations made by \cite{Yoon_GAIN_2018} (Figure 2).

\bibliography{references}